\documentclass[11pt,letterpaper]{article} 

\usepackage[OT1]{fontenc} 
\usepackage[numbers]{natbib}

\usepackage{fullpage,times}
\usepackage{booktabs}
\usepackage{amsthm,amsmath,amsfonts} 
\usepackage{epsf,epsfig,caption,subcaption,graphicx} 
\usepackage[ruled,boxed]{algorithm2e}
\usepackage{tikz}
\usepackage[colorlinks]{hyperref}

\newtheorem{theorem}{Theorem}[section]
 
\newtheorem{proposition}[theorem]{Proposition} 

\theoremstyle{definition}
\newtheorem{definition}[theorem]{Definition} 
 
\newtheorem{assumption}[theorem]{Assumption}

\theoremstyle{remark}

\def\var{\mbox{Var}} 

\def\E{\mathbb{E}} 
\def\S{\widehat{S}}
\def\P{P}

\def\exp{\mbox{exp}}
\def\pa{\mbox{Pa}}

\def\de{\mbox{De}}
\def\nd{\mbox{Nd}}

\DeclareMathAlphabet\mathbfcal{OMS}{cmsy}{b}{n}

\begin{document}

\begin{center}
	{\bf{\LARGE{Identifiability of Generalized Hypergeometric Distribution (GHD) Directed Acyclic Graphical Models}}}
	
	\vspace*{.1in}
	\begin{tabular}{c c}
		Gunwoong Park$^1$ & Hyewon Park$^1$
	\end{tabular}
	
	\vspace*{.1in}
	
	\begin{tabular}{c}
		$^1$ Department of Statistics, University of Seoul \\
	\end{tabular}
	
	\vspace*{.1in}
	

\end{center}


\begin{abstract}
We introduce a new class of identifiable DAG models where the conditional distribution of each node given its parents belongs to a family of generalized hypergeometric distributions (GHD). A family of generalized hypergeometric distributions includes a lot of discrete distributions such as the binomial, Beta-binomial, negative binomial, Poisson, hyper-Poisson, and many more. We prove that if the data drawn from the new class of DAG models, one can fully identify the graph structure. We further present a reliable and polynomial-time algorithm that recovers the graph from finitely many data. We show through theoretical results and numerical experiments that our algorithm is statistically consistent in high-dimensional settings ($p >n$) if the indegree of the graph is bounded, and out-performs state-of-the-art DAG learning algorithms.
\end{abstract}


\section{Introduction}
Probabilistic directed acyclic graphical (DAG) models or Bayesian networks provide a widely used framework for representing causal or directional dependence relationships among many variables. One of the fundamental problems associated with DAG models is learning a large-scale causal structure given samples from the joint distribution $\P(G)$ over a set of nodes of a graph $G$.

Prior works have addressed the question of identifiability for different classes of joint distribution $\P(G)$. \cite{frydenberg1990chain,heckerman1995learning} show the Markov equivalence class (MEC) where graphs that belong to the same MEC have the same conditional independence relations. \cite{chickering2003optimal, spirtes2000causation, tsamardinos2003towards, zhang2016three} show that the underlying graph of a DAG model is recoverable up to MEC under the faithfulness or some related conditions. However since many MECs contain more than one graph, a true graph cannot be determined. 

Recently, many works show fully identifiable DAG models under stronger assumptions on joint distribution $\P(G)$. \cite{peters2014identifiability} proves that Gaussian structural equation models with equal or known error variances are identifiable. In addition, \cite{shimizu2006linear} shows that linear non-Gaussian models where each variable is determined by a linear function of its parents plus a non-Gaussian error term are identifiable. \cite{hoyer2009nonlinear, mooij2009regression, peters2012identifiability} relax the assumption of linearity and prove that nonlinear additive noise models where each variable is determined by a non-linear function of its parents plus an error term are identifiable under suitable regularity conditions. Instead of considering linear or additive noise models, \cite{park2015learning, park2017learning} introduce discrete DAG models where the conditional distribution of each node given its parents belongs to the exponential family of discrete distributions such as Poisson, binomial, and negative binomial. They prove that the discrete DAG models are identifiable as long as the variance is a quadratic function of the mean. 


Learning DAG or causal discovery from \emph{count data} is an important research problem because such count data are increasingly ubiquitous in big-data settings, including high-throughput genomic sequencing data, spatial incidence data, sports science data, and disease incidence data. However as we discussed, most existing methods focus on the continuous or limited discrete DAG models. Hence it is important to model complex multivariate count data using a broader family of discrete distributions. 

In this paper, we generalize the main idea in \cite{park2015learning, park2017learning} to a \textit{family of generalized hypergeometric distributions (GHD)} that includes Poisson, hyper-Poisson, binomial, negative binomial, beta-binomial, hypergeometric, inverse hypergeometric and many more (see more examples in Table~\ref{table:example1} and  \cite{dacey1972family, kemp1968wide, kemp1974family}). We introduce a new class of identifiable DAG models where the conditional distribution of each node given its parents belongs to a family of GHDs. In addition, we prove that the class of GHD DAG models is identifiable from the joint distribution $\P(G)$ using \emph{convex} relationship between the mean and the r-th factorial moment for some positive integer $r$ under the causal sufficiency assumption that all relevant variables have been observed. However we do not assume the faithfulness assumption that can be very restrictive \cite{uhler2013geometry}. 

We also develop the reliable and scalable Moments Ratio Scoring (MRS) algorithm which learns any large-scale GHD DAG model. We provide computational complexity and statistical guarantees of our MRS algorithm to show that it has polynomial run-time and is consistent for learning GHD DAG models, even in the high-dimensional $p > n$ setting when the indegree of the graph $d$ is bounded. We demonstrate through simulations and a real NBA data that our MRS algorithm performs better than state-of-the-art GES \cite{chickering2003optimal}, MMHC \cite{tsamardinos2006max}, and ODS \cite{park2015learning} algorithms in terms of both run-time and recovering a graph structure. 

The remainder of this paper is structured as follows: Section 2.1 summarizes the necessary notation, Section 2.2 defines GHD DAG models and Section 2.3 proves that GHD DAG models are identifiable. In Section 3, we develop a polynomial-time algorithm for learning GHD DAG models and provide its theoretical guarantees and computational complexity in terms of the triple $(n,p,d)$. Section 4 empirically evaluates our methods compared to GES, MMHC, and ODS algorithms on synthetic and real basketball data. 


\section{GHD DAG Models and Identifiability}

\label{SecClass}

In this section, we first introduce some necessary notations and definitions for directed acyclic graph (DAG) models. Then we propose novel generalized hypergeometric distribution (GHD) DAG models for multivariate count data. Lastly, we discuss their identifiability using a convex relation between the mean and r-th factorial moments.


\subsection{Problem set-up and notation}

A DAG $G = (V, E)$ consists of a set of nodes $V = \{1, 2, \cdots, p\}$ and a set of directed edges $E \in V \times V$ with no directed cycles. A directed edge from node $j$ to $k$ is denoted by $(j,k)$ or $j \rightarrow k$. The set of \emph{parents} of node $k$ denoted by $\pa(k)$ consists of all nodes $j$ such that $(j,k) \in E$. If there is a directed path $j\to \cdots \to k$, then $k$ is called a \emph{descendant} of $j$ and $j$ is an \emph{ancestor} of $k$. The set $\de(k)$ denotes the set of all descendants of node $k$. The \emph{non-descendants} of node $k$ are $\nd(k) := V \setminus (\{k\} \cup \de(k))$. An important property of DAGs is that there exists a (possibly non-unique) \emph{ordering} $\pi = (\pi_1, ...., \pi_p)$ of a directed graph that represents directions of edges such that for every directed edge $(j, k) \in E$, $j$ comes before $k$ in the ordering. Hence learning a graph is equivalent to learning an ordering and skeleton that is a set of edges without their directions.

We consider a set of random variables $X := (X_j)_{j \in V}$ with a probability distribution taking values in probability space $\mathcal{X}_{v}$ over the nodes in $G$. Suppose that a random vector $X$ has a joint probability density function $P(G) = P(X_1, X_2, ..., X_p)$. For any subset $S$ of $V$, let $X_{S} :=\{X_j : j \in S \subset V \}$ and $\mathcal{X}(S) := \times_{j \in S} \mathcal{X}_{j}$. For any node $j \in V$, $\P(X_j \mid X_{S})$ denotes the conditional distribution of a variable $X_j$ given a random vector $X_{S}$. Then, a DAG model has the following factorization \cite{lauritzen1996graphical}: 
\begin{equation*}
\P(G) = \P(X_1, X_2,..., X_p) = \prod_{j=1}^{p} \P(X_j \mid X_{\pa(j)}),  
\end{equation*}
where $\P(X_j \mid X_{\pa(j)})$ is the conditional distribution of a variable $X_j$ given its parents $X_{\pa(j)}$. 

We suppose that there are $n$ i.i.d samples $X^{1:n} := ( X^{(i)} )_{i=1}^{n}$ drawn from a given DAG models where $X^{(i)} := (X_1^{(i)}, X_2^{(i)},\cdots, X_p^{(i)})$ is a $p$-variate random vector. We use the notation $~\widehat{\cdot}~$ to denote an estimate based on samples $X^{1:n}$. In addition, we assume the causal sufficiency that all variables have been observed.


\subsection{Generalized Hypergeometric Distribution (GHD) DAG Models}


In this section, we begin by introducing a family of generalized hypergeometric distributions (GHDs) defined by \cite{kemp1968wide}. A family of GHDs includes a large number of discrete distributions and has a special form of probability generating functions expressed in terms of the generalized hypergeometric series. We borrow the notations and terminologies in \cite{kemp1974family} to explain detailed properties of a family of GHDs. 
Let $\langle a \rangle ^j = a(a+1) \cdots (a + j -1)$ be the rising factorial, $(a)_j = a(a-1) \cdots (a - j +1)$ be the falling factorial, and $\langle a \rangle^{0} =(a)_{0} =1$. In addition, generalized hypergeometric function is:
\begin{equation*}
_pF_q[a_1,...,a_p; b_1,...,b_q; \theta] := \sum_{j \geq 0} \frac{ \langle a_1 \rangle ^j \cdots \langle a_p \rangle ^j \theta^j }{  \langle b_1\rangle^j \cdots \langle b_q\rangle^j j! }.
\end{equation*}

\cite{kemp1968wide, kemp1974family} show that GHDs have probability generating functions of the following form:
\begin{eqnarray*}
\label{eq:pgf}
G(s) = \frac{ _pF_q[a_1,...,a_p; b_1,...,b_q; \theta s]  }{_pF_q[a_1,...,a_p; b_1,...,b_q; \theta] } = 
~ _p F_q[a_1,...,a_p; b_1,...,b_q; \theta (s-1)] 
\end{eqnarray*}

This class of distributions includes a lot of discrete distributions such as binomial, beta-binomial, Poisson, Poisson type, displaced Poisson, hyper-Poisson, logarithmic, and generalized log-series. We provide more examples with their probability generating functions in Table~\ref{table:example1} (see also in \cite{dacey1972family,kemp1968wide, kemp1974family}). 
\begin{table}
	\begin{tabular}{lll}
		Distributions & p.g.f. $G(s) $ & Parameters\\ \hline 
		Poisson &  $_0F_0[; ; \lambda(s-1)]$ & $\lambda > 0$ \\
		Hyper-Poisson (Bardwell and Crow) & $_1F_1[1; b; \lambda(s-1)]$ & $\lambda > 0$ \\
		Negative Binomial & $_1F_0[k; ; p(s-1)]$ & $k, p > 0$ \\
		Poisson Beta & $_1F_1[a; a+b; \lambda(s-1)]$ & $a, b, \lambda >0$ \\
		Negative Binomial Beta & $_2F_1[k, a; a+b; \lambda(s-1)]$ & $k, a, b, \lambda >0$ \\
		STERRED Geometric & $_2F_1[1, 1; 2; q (s-1)/(1-q)]$ & $ 1> q > 0$ \\
		Shifted UNSTERRED Poisson & $_1F_1[2; 1; \lambda(s-1)]$ & $ 1\geq \lambda > 0$ \\
		Shifted UNSTERRED Negative Binomial & $_2F_1[k,2; 1; p(s-1)]$ & $ p>0, (p+1)/p \geq k > 0$  \\ \hline
	\end{tabular}
	\caption{Examples of Hypergeometric Distribution and their probability generating functions $G(s)$}
	\label{table:example1}
\end{table}

Now we define the generalized hypergeometric distribution (GHD) DAG models: 
\begin{definition}[GHD DAG Models]
	\label{def:CMRDAG}
	The DAG models belong to generalized hypergeometric distribution (GHD) DAG models if the conditional distribution of each node given its parents belongs to a family of generalized hypergeometric distributions and the parameter depend only on its parents: For each $j \in V$, $X_j \mid X_{\pa(j)}$ has the following probability generating function
	$$
	G\big(s  ; a(j), b(j) \big) = _{p_j}\!\!F_{q_j} [a(j); b(j); \theta(X_{\pa(j)}) (s-1)]
	$$
	where $a(j) = (a_{j1},..., a_{j p_j})$, $b(j) = (b_{j1},..., b_{j q_j})$, and $\theta: \mathcal{X}_{\pa(j)} \to \mathbb{R}$.
\end{definition}
A popular example of GHD DAG models is a Poisson DAG model in \cite{park2015learning} where a conditional distribution of each node $j \in V$ given its parents is Poisson and the rate parameter is an arbitrary positive function $\theta_j(X_{\pa(j)})$. Unlike Poisson DAG models, GHD DAG models are hybrid models where the conditional distributions have various distributions which incorporate different data types. In addition, the exponential family of discrete distributions discussed in \cite{park2017learning} is also included in a family of GHDs. Hence, our class of DAG models is strictly broader than the previously studied identifiable DAG models for multivariate count data.

GHD DAG models have a lot of useful properties for identifying a graph structure. One of the useful properties is the recurrence relation involving factorial moments:
\begin{proposition}[Constant Moments Ratio (CMR) Property]
	\label{prop:factorial}
	Consider a GHD DAG model. Then for any $j \in V$ and any integer $r = 2,3,...$, there exists a \emph{r-th factorial constant moments ratio (CMR) function} $f_{j}^{(r)}(x ;  a(j), b(j) ) = x^r \prod_{i = 1}^{p_j} \left( \frac{ (a_{ji}+r-1)_r }{ a_{ji}^r } \right) \prod_{k = 1}^{q_j} \left( \frac{ b_{jk}^r }{ (b_{jk}+r-1)_r  } \right)$ such that 
	\begin{align*}
	\E\big( (X_j)_r \mid X_{\pa(j)} \big) 
	= f_{j}^{(r)} \big( \E(X_j \mid X_{\pa(j)} ) ;  a(j), b(j) \big).
	\end{align*}
	as long as $\max X_j \geq r$.
\end{proposition}
The detail of the proof is provided in Appendix. Prop.~\ref{prop:factorial} claims that the GHD DAG models always satisfy the \emph{r-th constant moments ratio (CMR) property} that the r-th factorial moment is a function of the mean. The condition $\max X_j \geq r$ for $r \geq 2$ rules out the DAG models with Bernoulli and multinomial distributions which are known to be non-identifiable \cite{heckerman1995learning}. We will exploit the CMR property for model identifiability in the next section.

\subsection{Identifiability}

\label{SecIden}

\begin{figure}[!t]
	\centering
	\begin {tikzpicture}[ -latex ,auto,
	state/.style={circle, draw=black, fill= white, thick, minimum size= 2mm},
	label/.style={thick, minimum size= 2mm}
	]
	\node[state] (X1)  at (0,0)   {\small{$X_1$} }; \node[state] (X2)  at (2,0)   {\small{$X_2$}}; \node[label] (X3) at (1,-.7) {$G_1$};
	\node[state] (Y1)  at (5,0)   {\small{$X_1$}}; \node[state] (Y2)  at (7,0)   {\small{$X_2$}}; \node[label] (Y3) at (6,-.7) {$G_2$};
	\node[state] (Z1)  at (10,0)   {\small{$X_1$}}; \node[state] (Z2)  at (12,0)   {\small{$X_2$}}; \node[label] (Z3) at (11,-.7) {$G_3$};
	\path (Y1) edge [shorten <= 2pt, shorten >= 2pt] node[above]  { } (Y2); 
	\path (Z2) edge [shorten <= 2pt, shorten >= 2pt] node[above]  { } (Z1);
\end{tikzpicture}
\caption{Bivariate directed acyclic graphs of $G_1$, $G_2$ and $G_3$  }
\label{figure1}
\end{figure}
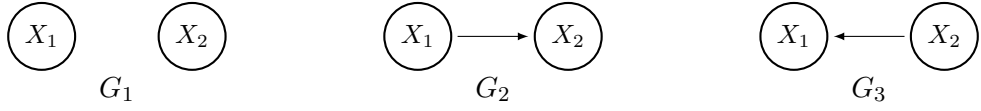

In this section we prove that GHD DAG models are identifiable. To provide intuition, we show identifiability for the bivariate Poisson DAG model discussed in \cite{park2015learning}. Consider all possible graphical models illustrated in Fig.~\ref{figure1}: $G_1: X_1 \sim \mbox{Poisson}(\lambda_1),\;\; X_2 \sim \mbox{Poisson}(\lambda_2)$, where $X_1$ and $X_2$ are independent; $G_2: X_1 \sim \mbox{Poisson}(\lambda_1)$ and $X_2 \mid X_1 \sim \mbox{Poisson}(\theta_2(X_1))$; and $G_3: X_2 \sim \mbox{Poisson}(\lambda_2)$ and $X_1 \mid X_2 \sim \mbox{Poisson}(\theta_1(X_2))$ for arbitrary positive functions $\theta_1, \theta_2: \mathbb{N} \cup \{0\} \to \mathbb{R}^{+}$. Our goal is to determine whether the underlying graph is $G_1, G_2$ or $G_3$ from  the probability distribution $P(G)$.

We exploit the CMR property for Poisson, $\E( (X_j)_r ) = \E(X_j)^r$ for any positive integer $r \in \{2,3,...\}$. For $G_1$, $\E( (X_1)_r )= \E(X_1)^r$ and $\E( (X_2)_r )= \E(X_2)^r$. For $G_2$, $\E( (X_1)_r )= \E(X_1)^r$, while
\begin{align*}
\E( (X_2)_r ) &= \E( \E( (X_2)_r \mid X_1) ) = \E( \E(X_2 \mid X_1)^r ) > \E( \E(X_2 \mid X_1) )^r = \E(X_2)^r,
\end{align*}
as long as $\E(X_2 \mid X_1)$ is not a constant. The inequality follows from the Jensen's inequality.

Similarly for $G_3$, $\E( (X_2)_r )= \E(X_2)^r$ and $\E( (X_1)_r ) >  \E(X_1)^r$ as long as $\E(X_1 \mid X_2)$ is not a constant. Hence we can distinguish graphs $G_1$, $G_2$, and $G_3$ by testing whether a moments ratio $\E( (X_j)_r ) / \E(X_j)^r$ is greater than or equal to 1. 

Now we state the identifiability condition for the general case of p-variate GHD DAG models:

\begin{assumption}[Identifiability Condition]
\label{Ass1}
For a given GHD DAG model, the conditional distribution of each node given its parents is known. In other words, the r-th factorial CMR functions $(f_j^{(r)}(x ; a(j), b(j) ) )_{j \in V}$ are known. Moreover, for any node $j \in V$, $\E(X_j \mid X_{\pa(j)})$ is non-degenerated. 
\end{assumption}
Prop.~\ref{prop:factorial} and Assumption~\ref{Ass1} enable us to use the following property: for any node $j \in V$, $\E( (X_j)_r ) = \E( f_j^{(r)}( \E( X_j  \mid X_{\pa(j)} ) ; a(j), b(j) ) )$, while for any non-empty $\pa_0(j) \subset \pa(j)$ and $S_j \subset  \nd(j) \setminus \pa_0(j)$, 
\begin{align*}
\E( (X_j)_r ) &= \E( \E( f_j^{(r)} ( \E( X_j \mid X_{\pa(j)}) ; a(j), b(j) ) \mid X_{S_j} ) )  > \E( f_j^{(r)}( \E( X_j  \mid X_{S_j} ) ; a(j), b(j) ) ),
\end{align*}
because the CMR function is strictly convex. 

We state the first main result that general $p$-variate GHD DAG models are identifiable: 
\begin{theorem}[Identifiability]
\label{Thm1}
Under Assumption~\ref{Ass1}, the class of GHD DAG models is identifiable.
\end{theorem}
We defer the proof in Appendix. The key idea of the identifiability is to search a smallest conditioning set $S_j$ for each node $j$ such that the moments ratio $\E( (X_j)_r ) / \E( f_j^{(r)}( \E( X_j \mid X_{S_j} ) ) ) =  1$. Thm.~\ref{Thm1} claims that the assumption on nodes distributions is sufficient to uniquely identify GHD DAG models. In other words, the well-known assumptions such as faithfulness, non-linear causal relation, non-Gaussian additive noise assumptions are not necessary \cite{hoyer2009nonlinear, mooij2009regression, peters2014identifiability, peters2012identifiability, shimizu2006linear}. 

Thm.~\ref{Thm1} implies that Poisson DAG models are identifiable even when the form of rate parameter functions $\theta_j$ are unknown. Thm.~\ref{Thm1} also claims that hybrid DAG models, in which the distributions of nodes are different, are identifiable as long as the distributions are known while the forms of parameter functions are unknown. In Section~\ref{SecNum}, we provide numerical experiments on Poisson and hybrid DAG models to support Thm.~\ref{Thm1}.
  
\section{Algorithm}

\label{SecAlg}

In this section, we present our Moments Ratio Scoring (MRS) algorithm for learning GHD DAG models. Our MRS algorithm has two main steps: 1) identifying the skeleton (i.e., edges without their directions) using existing skeleton learning algorithms; and 2) estimating the ordering of the DAG using moments ratio scores, and assign the directions to the estimated skeleton based on the estimated ordering.

Although GHD DAG models can be recovered only using the r-th CMR property according to Thm.~\ref{Thm1}, our algorithm exploits the skeleton to reduce the search space of DAGs. From the idea of constraining the search, our algorithm achieves computational and statistical improvements. More precisely, Step 1) provides \emph{candidate parents} set for each node. The concept of candidate parents set exploits two properties; (i) the neighborhood of a node $j$ in the graph denoted by $\mathcal{N}(j):= \{k \in V \mid (j,k)$ or $(k,j) \in E \}$ is a superset of its parents, and (ii) a node $j$ should appear later than its parents in the ordering. Hence, the candidate parents set for a node $j$ is the intersection of its neighborhood and elements of the ordering which appear before that node $j$, and is denoted by $C_{m j} := \mathcal{N}(j) \cap \{\pi_1, \pi_2,...,\pi_{m-1} \}$ where $m^{th}$ element of the ordering is $j$ (i.e., $\pi_{m} = j$). The estimated candidate parents set is $\widehat{C}_{m j} := \widehat{\mathcal{N}}(j) \cap \{\widehat{\pi}_1, \widehat{\pi}_2,...,\widehat{\pi}_{m-1} \}$ that is specified in Alg.\ref{AlgMRS}.

This candidate parents set is used as a conditioning set for a moments ratio score in Step 2). If the candidate parents set is not applied, the size of the conditioning set for a  moments ratio score could be $p-1$. Since Step 2) computes the r-th factorial moments, the sample complexities depends significantly on the number of variables we condition on as illustrated in Sections \ref{SecStat}. Therefore by making the conditioning set for a moments ratio score of each node as small as possible, we gain huge statistical improvements.

The idea of reducing the search space of DAGs has been studied in many sparse candidate algorithms \cite{zhang2009identifiability, hyvarinen2013pairwise}. Hence for Step 1) of our algorithm, any off-the-shelf candidate parents set learning algorithms can be applied such as the MMPC~\citep{tsamardinos2003towards} algorithm. Moreover, any standard MEC learning algorithms such as GES~\citep{chickering2003optimal}, PC~\citep{spirtes2000causation} and MMHC~\cite{tsamardinos2006max} can be exploited because MEC provides the skeleton of a graph~\cite{verma1990equivalence}. In Section~\ref{SecNum}, we provide the simulation results of the MRS algorithm where GES and MMHC algorithms are applied in Step 1).

Step 2) of the MRS algorithm involves learning the ordering by comparing moments ratio scores of nodes using~Eqn. \eqref{EqnTruncScorej}. The basic idea is to test which nodes have moment ratio score $1$. The ordering is determined one node at a time by selecting the node with the smallest moment ratio score because the correct element of the ordering has the score 1, otherwise strictly greater than 1 in population. 

Regarding the moments ratio scores, the score can be exploited for recovering the ordering only if the CMR property holds, which implies that the score should not be zero. Even if the zero value score is impossible in population, zero value scores often arise for a low count data such that all samples are less than $r$. Hence in order to avoid zero value scores due to a sample r-th factorial moment (i.e., $\widehat{\E}( (X)_r ) = 0$), we use an alternative ratio $\E( X^r ) / \big( f^{(r)}( \E(X) )-\sum_{k=0}^{r-1} s(r,k) \E(X^k) \big)$ where $s(r,k)$ is Stirling numbers of the first kind. This alternative ratio score comes from $(x)_r = \sum_{k=0}^{r} s(r,k) x^k$, therefore $\E( X^r ) = f^{(r)}( \E(X) ) -\sum_{k=0}^{r-1} s(r,k) \E(X^k)$. 

Hence the moments ratio scores in Step 2) of Alg.\ref{AlgMRS} involve the following equations:  \vspace{-2mm}
\begin{align}
\label{EqnTruncScorej}
\widehat{\mathcal{S}}_r(1,j) :=& 
\frac{ \widehat{ \E} ( X_j^r ) }{f_j^{(r)} ( \widehat{\E}( X_j ) )  -\sum_{k=0}^{r-1} s(r,k) \widehat{\E}(X_j^k) } \text{ , and} \\
\widehat{\mathcal{S}}_r(m,j) :=&  
\sum_{x \in \mathcal{X}_{\widehat{C}_{mj}} } \frac{ n_{\widehat{C}_{mj}}(x) }{n_{\widehat{C}_{mj}}} \left[  \frac{ \widehat{\E}(X_j^r\mid X_{\widehat{C}_{mj}} = x  ) }{ f_j^{(r)} (\widehat{\E}(X_j\mid X_{\widehat{C}_{mj}} = x  )) - \sum_{k=0}^{r-1} s(r,k) \widehat{\E}(X_j^r\mid X_{\widehat{C}_{mj}} = x  ) }\right]  \nonumber .
\end{align}
where $\widehat{C}_{mj}$ is the estimated candidate parents set of node $j$ for the $m^{th}$ element of the ordering. In addition, $n(x_S) := \sum_{i=1}^n \mathbf{1}( X_S^{(i)} = x_S )$ if $n(x_S) \geq N_{\min}$ otherwise 0, that refers to the truncated conditional sample size for $x_S$, and $n_S := \sum_{ x_S } n(x_S)$ refers to the total truncated conditional sample size for variables $X_S$. We discuss the choice of $N_{\min}$ later in Sec~\ref{SecStat}. Lastly, we use the method of moments estimators $\widehat{\E}( X_j^k ) = \frac{1}{n} \sum_{i = 1}^{n} ( (X_j^{(i)})^k )$ as unbiased estimators for $\E( X_j^k )$ for $1 \leq k \leq r$.

Since there are many conditional distributions, our moments ratio score is the weighted average of the levels of how well each distribution satisfies the r-th CMR property. 
The score only contains the conditional expectations with $n(x_S) \geq N_{\min}$ for better accuracy because the accuracy of the estimation of a conditional expectation $\widehat{E}(X_j \mid  x_S)$ relies on the sample size.

Finally, a directed graph is estimated combining the estimated skeleton from Step 1) and the estimated ordering from Step 2) that is 
$\widehat{E} :=  \cup_{j \in V} \{ (k, \widehat{\pi}_j) \mid k\in \widehat{\mathcal{N}}(\widehat{\pi}_j) \cap ( \widehat{\pi}_1,\widehat{\pi}_2,...,\widehat{\pi}_{j-1} ) \} $.
\setlength{\algomargin}{0.5em}
\begin{algorithm}[!t]
	\caption{ \bf Moments Ratio Scoring~\label{AlgMRS} }
	\SetKwInOut{Input}{Input}
	\SetKwInOut{Output}{Output}
	\SetKwInOut{Return}{Return}
	\Input{$n$ i.i.d. samples, $X^{1:n}$}
	\Output{Estimated ordering $\widehat{\pi}$ and an edge structure, $\widehat{E} \in V \times V$ }
	\BlankLine
	Step 1: Estimate the skeleton of the graph $\widehat{\mathcal{N}}$\;
	Step 2: Estimate an ordering of the graph using r-th moments ratio scores\; 
	Set $\widehat{\pi}_{0} = \emptyset$\;
	\For{$m = \{1,2,\cdots,p-1\}$}{
		\For{$j \in \{1,2,\cdots,p\} \setminus \{\widehat{\pi}_1,\cdots,\widehat{\pi}_{m-1}\}$ }{
			Find candidate parents set $\widehat{C}_{m j} = \widehat{\mathcal{N}}(j) \cap \{\widehat{\pi}_1, \cdots,\widehat{\pi}_{m-1}\}$\;
			Calculate r-th moments ratio scores $\widehat{\mathcal{S}}_r(m,j)$ using Eqn. \eqref{EqnTruncScorej};
		}
		The $m^{th}$ element of the ordering $\widehat{\pi}_m = \arg \min_j \widehat{\mathcal{S}}(m,j)$\;
	}
	The last element of the ordering $\widehat{\pi}_p = \{1,2,\cdots,p\} \setminus \{\widehat{\pi}_1, \widehat{\pi}_2,\cdots, \widehat{\pi}_{p-1}\}$\;
	\Return{Estimate the edge sets: $\widehat{E} =  \cup_{m \in V} \{ (k, \widehat{\pi}_m) \mid k\in \widehat{\mathcal{N}}(\widehat{\pi}_m) \cap ( \widehat{\pi}_1,...,\widehat{\pi}_{m-1} )  \} $}
	
\end{algorithm} 

\subsection{Computational Complexity}

\label{SecComp}

The MRS algorithm uses any skeleton learning algorithms with known computational complexity for Step 1). Hence we first focus on our novel Step 2) of the MRS algorithm. In Step 2), there are $(p-1)$ iterations and each iteration has a number of moments ratio scores to be computed which is bounded by $O(p)$. Hence the total number of scores to be calculated is $O(p^2)$. The computation time of each score is proportional to the sample size $n$, the complexity is $O( n p^2 )$. 

The total computational complexity of the MRS algorithm depends on the choice of the algorithm in Step 1). Since learning a DAG model is NP-hard \cite{chickering1994learning}, many state-of-the-art DAG learning algorithms such as GES \cite{chickering2003optimal},  GDS \cite{peters2014identifiability, peters2011causal}, PC \cite{spirtes2000causation}, and  MMHC \cite{tsamardinos2006max} are inherently heuristic algorithms. Although these algorithms take greedy search strategies, the computational complexities of greedy search based GES and MMHC algorithms are empirically $O(n^2 p^2)$. In addition, PC algorithm runs in the worst case in exponential time. Hence, Step 2) may not the main computational bottleneck of the MRS algorithm. In Section~\ref{SecNum}, we compare the MRS to GES algorithm in terms of log run-time, and show that the addition of estimation of ordering does not significantly add to the computational bottleneck.

\subsection{Statistical Guarantees}

\label{SecStat}

The MRS algorithm exploits well-studied existing algorithms for Step 1). Hence, we focus on theoretical guarantees for Step 2) of the MRS algorithm given that the skeleton is correctly estimated in Step 1). The main result is expressed in terms of the triple $(n,p,d)$ where $n$ is a sample size, $p$ is a graph node size, and $d$ is the indegree of a graph. Lastly, we discuss the sufficient conditions for recovering the graph via the MRS algorithm according to the chosen skeleton learning algorithm for Step 1).

We begin by discussing three required conditions that  the MRS algorithm recovers the ordering of a graph.
\begin{assumption}
	\label{A1}
	Consider the class of GHD DAG models with r-th factorial CMR function $f_j^{(r)}$ specified in Prop.~\ref{prop:factorial}. For all $j \in V$, any non-empty $\pa_0(j) \subset \pa(j)$, and $S_j \subset  \nd(j) \setminus \pa_0(j)$,
	\begin{itemize} \vspace{-1mm}
		\item[(A1)] there exists a positive constant $M_{\min} > 0$ such that 
		$$
		 \frac{ \widehat{\E}(X_j^r\mid X_{S_j}) }{ f_j^{(r)} (\widehat{\E}(X_j\mid X_{S_j})) - \sum_{k=0}^{r-1} s(r,k) \widehat{\E}(X_j^r\mid X_{S_j}) } > 1 + M_{\min}
		$$
		
		\item[(A2)] there exists a positive constant $V_1$ such that
		$$
		\E( \exp( X_j ) \mid X_{\pa(j)} ) < V_1.  \vspace{-1mm}	
		$$
		
		\item[(A3)] there are some elements $x_{S_j} \in \mathcal{X}_{S_j}$ such that 
		$
		\sum_{i=1}^n \mathbf{1}( X_{S_j}^{(i)} = x_{S_j} ) \geq N_{\min}
		$
		where $N_{\min} > 0$ is the predefined minimum sample size in the MRS algorithm. 		
	\end{itemize}
\end{assumption} 

The first condition is a stronger version of Assumption~\ref{Ass1} since we move from the population to the finite sample setting. The second assumption is to control the tail behavior of the conditional distribution of each variable given its parents. It enables to control the accuracy of moments ratio scores in Eqn. \eqref{EqnTruncScorej} in high dimensional settings ($ p > n $). The last assumption ensures that the score can be calculated.


We now state the second main result under Assumption~\ref{A1}. Since the true ordering $\pi$ is possibly not unique, we use $\mathcal{E}(\pi)$ to denote the set of all the orderings that are consistent with the DAG $G$. 
\begin{theorem}[Recovery of the ordering]
	\label{ThmCausalOrdering}
	Consider a GHD DAG model where the conditional distribution of each node given its parents is known. Suppose that the skeleton of the graph is provided, the maximum indegree of the graph is $d$, and Assumptions~\ref{A1}(A1)-(A3) are satisfied.
	Then there exists constant $C_{\epsilon} > 0$ for any $\epsilon > 0$ such that if sample size is sufficiently large $n > C_{\epsilon} \log^{2r + d} (\max{(n, p)}) ( \log(p) + \log(r) )  $, the MRS algorithm with the $r$-th moments ratio scores recovers the ordering with high probability: 
	$$
	P( \widehat{\pi} \in \mathcal{E}(\pi) ) \geq 1 - \epsilon.
	$$
\end{theorem}
The detail of the proof is provided in Appendix. Intuitively, it makes sense because the method of moment estimator converges to the true moment as sample size $n$ increases by weak law of large number. This allows the algorithm to recover a true ordering for the DAG consistently. 

Thm.~\ref{ThmCausalOrdering} claims that if the sample size $n = \Omega( \log^{2r + d} (\max{(n, p)}) \log(p) )$, our MRS algorithm accurately estimates a true ordering with high probability. Hence our MRS algorithm works in high-dimensional settings ($p>n$) provided that the indegree of the graph $d$ is bounded. This theoretical result is also consistent with learning Poisson DAG models shown in~\cite{park2015learning} where if $n = \Omega( \log^{4 + d} (\max{(n, p)}) \log(p) )$ their ODS algorithm recovers the ordering well. Since \cite{park2015learning} uses the variance (the second order moments $r = 2$), both algorithms are expected to have the same performance of recovering graphs.

However the MRS algorithm performs significantly better than the ODS algorithm in general because the moments difference the ODS algorithm exploits is proportional to magnitude of the conditional mean while the moments ratio is not. For a simple Poisson DAG $X_1 \to X_2$, $\E( (X_2)_2) - \E(X_2)^2 = \var( E(X_2\mid X_1) )$. Hence if $\E(X_2\mid X_1) \approx 0$, the score in ODS is inevitably close to 0, while the score in MRS, $\E( (X_2)_2)/ \E(X_2)^2 = 1 + \var( \E(X_2\mid X_1) )/ \E( X_2 )^2$ is not necessarily close to 1. Hence, Assumption~\ref{A1}(A1) is much milder than the related assumption for the ODS algorithm. 

Now we discuss the sufficient conditions for recovering the true graph via the MRS algorithm according to the choice of the algorithm in Step 1). The  GES \cite{chickering2003optimal}, PC \cite{spirtes2000causation}, and MMHC \cite{tsamardinos2006max} algorithms require the Markov, faithfulness, and causal sufficiency or related assumptions to recover the skeleton of a graph. Moreover GES and MMHC algorithms are greedy search based algorithms that are not guaranteed to recover the true skeleton of a graph. Therefore, the MRS algorithm may require strong assumptions or large sample size to recover the true graph based on the choice of the algorithm in Step 1). Although these assumptions can be very restrictive, we show through the simulations that the MRS algorithm recovers the directed edges well even in high dimensional settings ($p>n$).


\section{Numerical Experiments}

\label{SecNum}

In this section, we provide various simulation results by comparing the MRS algorithm to state-of-the-art ODS, GES and MMHC algorithms in terms of recovering MECs and DAGs. Also, we support our theoretical results in Thm.~\ref{ThmCausalOrdering} and computational complexity in Section~\ref{SecComp} with synthetic and real basketball data. In addition, we show that our algorithm performs favorably compared to the ODS, GES, and MMHC algorithms in terms of recovering the directed graphs.

\subsection{Synthetic Data}

 We conduct two sets of simulation study using $150$ realizations of $p$-node random GHD DAG models with the indegree constraints $d =2$: (1) Poisson DAG models where the conditional distribution of each node given its parents is Poisson; and (2) Hybrid DAG models where the conditional distributions are sequentially Poisson, Binomial with $N = 3$, hyper-Poisson with $b= 2$, and Binomial with $N = 3$.

We set the (hyper) Poisson rate parameter $\theta_j(\pa(j)) = \exp( \theta_{j}+  \sum_{ k \in \pa(j) } \theta_{jk} X_k  )$ and the binomial probability $p_j(\pa(j)) = \mathrm{logit}^{-1}( \theta_{j} + \sum_{ k \in \pa(j) } \theta_{jk} X_k  )$. For Poisson DAG models, the set of non-zero parameters $\theta_{jk}\in \mathbb{R}$ were generated uniformly at random in the range $\theta_{jk} \in [-1.75, -0.25] \cup [0.25,~1.75]$ and $\theta_{j} \in [1, 3]$. For Hybrid DAG models, the $ \theta_{jk} $ were generated uniformly at random in range $\theta_{jk} \in [-1.2, -0.2]$ and $\theta_{j} \in [1, 3]$. These settings help the generated values of samples to avoid either all zeros (constant) or too large ($>10^{309}$). However if some samples are all zeros or too large, we regenerate parameters and samples. We also set the $r \in \{2,3,4\}$ and $N_{\min} = 1$ for computing the r-th moments ratio scores.


\begin{figure*}[!t] \vspace{-0mm}
	\centering \hspace{-2mm}
	\begin{subfigure}[!htb]{.24\textwidth}
		\includegraphics[width=\textwidth]{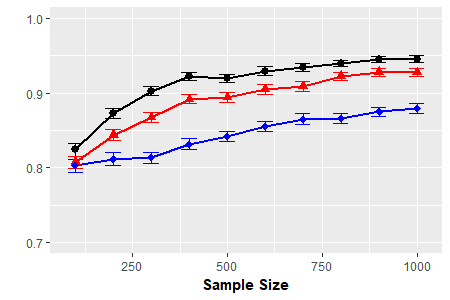}
		\caption{Poisson: $p=20$}
	\end{subfigure} \hspace{-2mm}
	\begin{subfigure}[!htb]{.24\textwidth}
		\includegraphics[width=\textwidth]{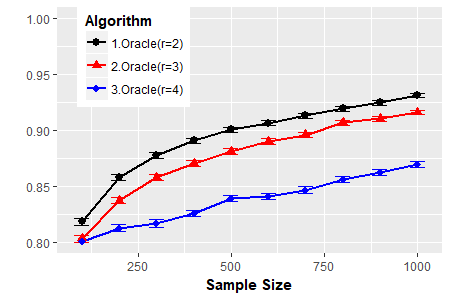}
		\caption{Poisson: $p=100$}
	\end{subfigure}	\hspace{-2mm}
	\begin{subfigure}[!htb]{.24\textwidth}
		\includegraphics[width=\textwidth]{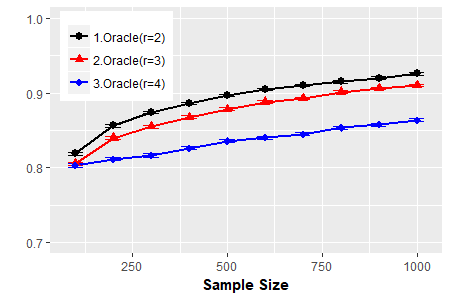}
		\caption{Poisson: $p=200$}
	\end{subfigure} \hspace{-2mm}
	\begin{subfigure}[!htb]{.24\textwidth}
		\includegraphics[width=\textwidth]{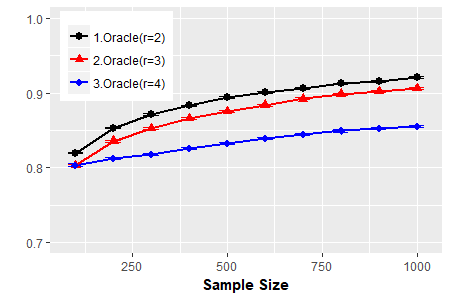}
		\caption{Poisson: $p=500$}
	\end{subfigure}	\hspace{-2mm}

\begin{subfigure}[!htb]{.24\textwidth}
	\includegraphics[width=\textwidth]{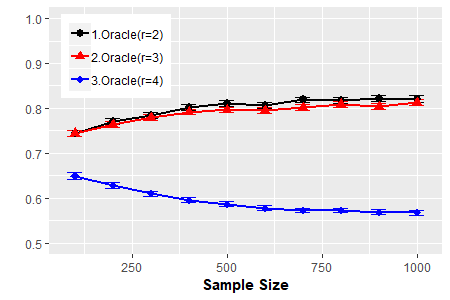}
	\caption{Hybrid: $p=20$}
\end{subfigure} \hspace{-2mm}
\begin{subfigure}[!htb]{.24\textwidth}
	\includegraphics[width=\textwidth]{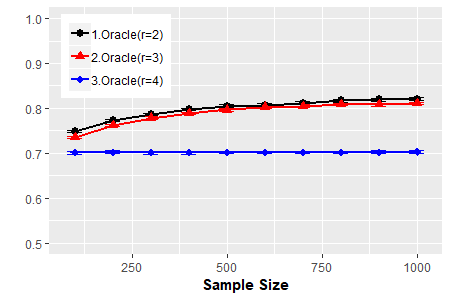}
	\caption{Hybrid: $p=100$}
\end{subfigure}	\hspace{-2mm}
	\begin{subfigure}[!htb]{.24\textwidth}
		\includegraphics[width=\textwidth]{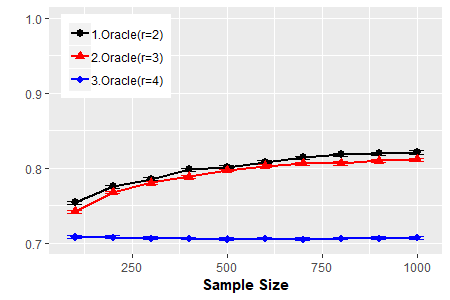}
		\caption{Hybrid: $p=200$}
	\end{subfigure}	\hspace{-2mm}
	\begin{subfigure}[!htb]{.24\textwidth}
		\includegraphics[width=\textwidth]{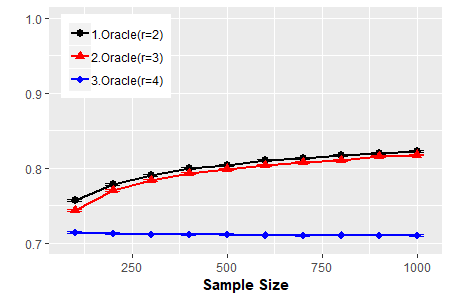}
		\caption{Hybrid: $p=500$}
	\end{subfigure}
	\caption{Comparison of the MRS algorithms using different values of $r = 2,3,4$ for the scores in terms of recovering the ordering of Poisson and Hybrid DAG models given the true skeletons.} 
	\label{fig:result001}
\end{figure*} 


\subsubsection{Recovery of the DAG}

In order to authenticate the validation of Thm.~\ref{ThmCausalOrdering}, we plot the average precision ($\tfrac{ \# \text{ of correctly estimated edges } }{ \# \text{ of estimated edges} }$) as a function of sample size $n \in \{100, 200,..., 1000\}$ for different node sizes $p = \{20, 100, 200, 500\}$ given the true skeleton. Fig.~\ref{fig:result001} provides a comparison of how accurately our MRS algorithm performs in terms of recovering the orderings of the GHD DAG models. Fig.~\ref{fig:result001} supports our main theoretical results in Thm.~\ref{ThmCausalOrdering}: (i) our algorithm recovers the ordering more accurately as sample size increases; (ii) our algorithm can recover the ordering in high dimensional settings; and (iii) the required sample size $n = \Omega( \log^{2r + d} (\max{(n, p)}) \log(p) )$ depends on the choice $r$ because our algorithm with $r = 2$ performs significantly better than our algorithms with $r = 3, 4$. For Hybrid DAG models with $r = 4$, the precision seems not to increase as sample size increases. It makes sense because Binomial with $N = 3$ cannot satisfy the CMR property~\ref{prop:factorial} and Assumption~\ref{A1} (A1) with $r = 4$ i.e., $\E( (X_j)_4 ) = 0$. However the precision 0.7 is significantly bigger than 0.5 which is the precision of the graph with a random ordering. 

\begin{figure*}[!t]
	\centering \hspace{-2mm}
	\begin{subfigure}[!htb]{.23\textwidth} 
		\includegraphics[width=\textwidth]{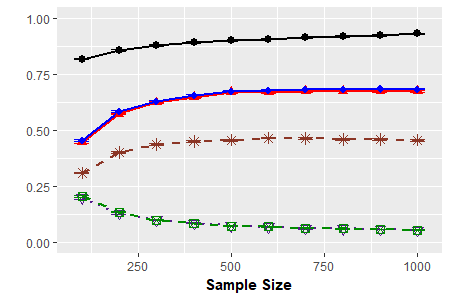}
		\caption{Precision: $p=100$ }
	\end{subfigure} \hspace{-2mm}
	\begin{subfigure}[!htb]{.23\textwidth}
		\includegraphics[width=\textwidth]{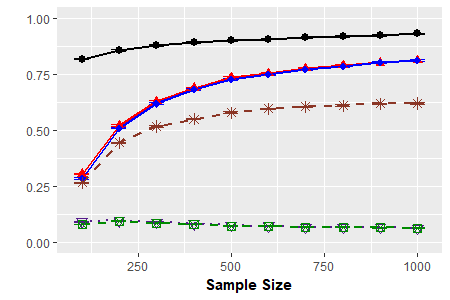}
		\caption{Recall: $p=100$}
	\end{subfigure} \hspace{-2mm}
	\begin{subfigure}[!htb]{.23\textwidth}
		\includegraphics[width=\textwidth]{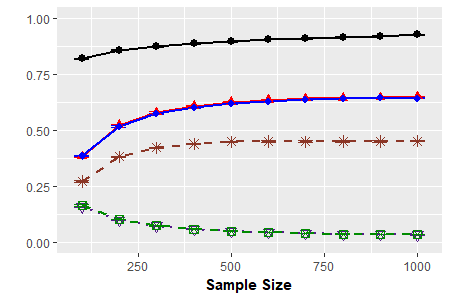}
		\caption{Precision: $p=200$}
	\end{subfigure} \hspace{-2mm}
	\begin{subfigure}[!htb]{.23\textwidth}
		\includegraphics[width=\textwidth]{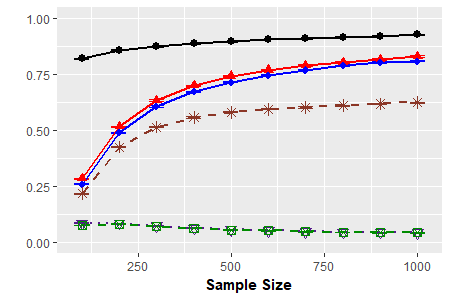}
		\caption{Recall: $p=200$}
	\end{subfigure}
	\begin{subfigure}[!htb]{.07\textwidth}
		\includegraphics[width=\textwidth, trim={13cm 0 0cm 1.3cm},clip]{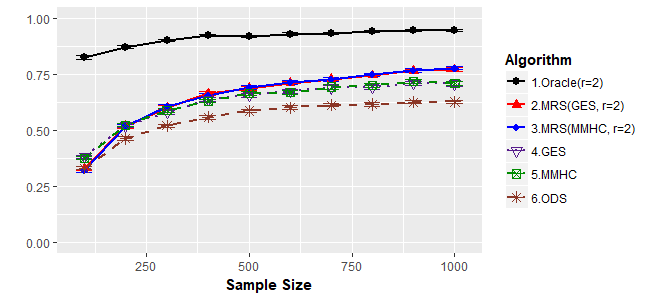}
	\end{subfigure}
\caption{Comparison of our MRS algorithms using GES and MMHC algorithms in Step 1) and $r = 2$ to the ODS, GES, MMHC algorithms in terms of recovering Poisson DAG models with $p \in \{100, 200\}$.  \vspace{-5mm} } 
\label{fig:result002P}
\end{figure*}

\begin{figure*}[!t]	
	\begin{subfigure}[!htb]{.23\textwidth} 
		\includegraphics[width=\textwidth]{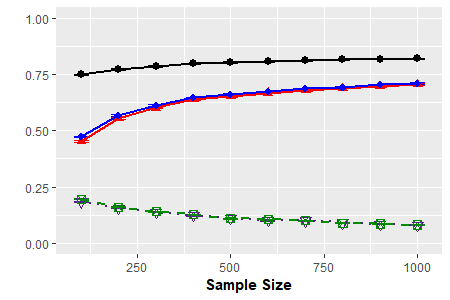}
		\caption{Precision: $p=100$}
	\end{subfigure} \hspace{-2mm}
	\begin{subfigure}[!htb]{.23\textwidth}
		\includegraphics[width=\textwidth]{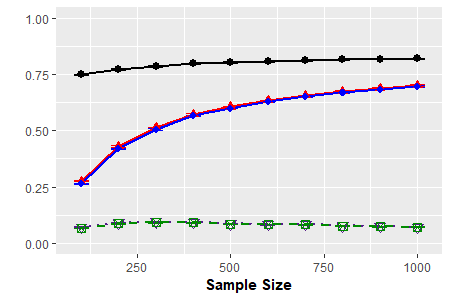}
		\caption{Recall: $p=100$}
	\end{subfigure} \hspace{-2mm}
	\begin{subfigure}[!htb]{.23\textwidth}
		\includegraphics[width=\textwidth]{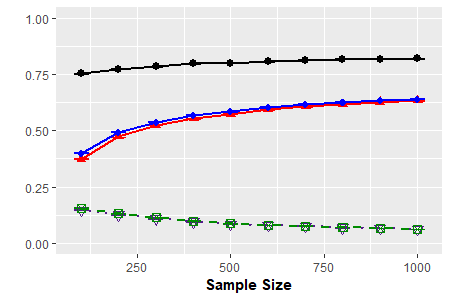}
		\caption{recision: $p=200$}
	\end{subfigure} \hspace{-2mm}
	\begin{subfigure}[!htb]{.23\textwidth}
		\includegraphics[width=\textwidth]{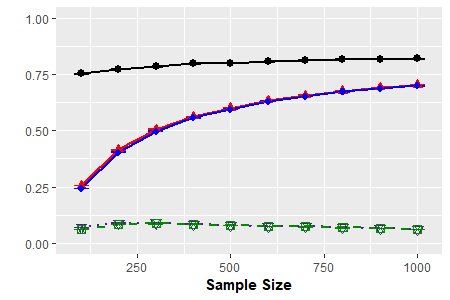}
		\caption{Recall: $p=200$}
	\end{subfigure}
	\begin{subfigure}[!htb]{.07\textwidth}
		\includegraphics[width=\textwidth, trim={14.2cm 0 0cm 1.3cm},clip]{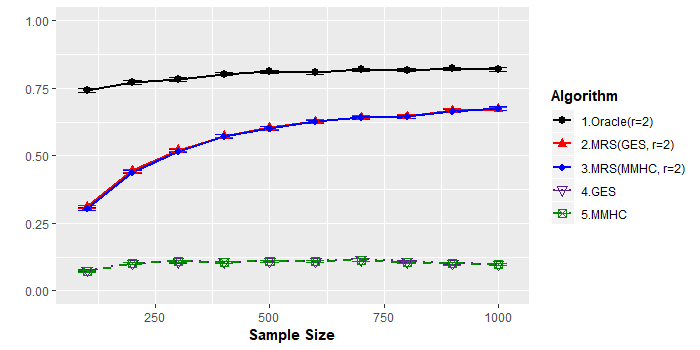}
	\end{subfigure}
	\caption{Comparison of our MRS algorithms using GES and MMHC algorithms in Step 1) and $r = 2$ to the GES, MMHC algorithms in terms of recovering Hybrid DAG models with $p \in \{100, 200\}$.  \vspace{-5mm} } 
	\label{fig:result002H}
\end{figure*}

In Figs.~\ref{fig:result002P} and \ref{fig:result002H}, we compare the MRS algorithm where $r = 2$ for the score, and GES and MMHC algorithms are applied in Step 1) to state-of-the art ODS, GES and MMHC algorithms by providing two results as a function of sample size $n \in \{100, 200,..., 1000\}$ for varying node size $p \in \{100, 200\}$: (i) the average precision ($\frac{ \# \text{ of correctly estimated edges } }{ \# \text{ of estimated edges} }$); (ii) the average recall ($\frac{ \# \text{ of correctly estimated edges } }{ \# \text{ of ture edges} }$). We also provide an oracle where the true skeleton is used while the ordering is estimated via the moments ratio scores.

As we see in Figs.~\ref{fig:result002P} and \ref{fig:result002H}, the MRS algorithm accurately recovers the true directed edges as sample size increases. However since the skeleton estimation is not perfect, we can see the performances of our MRS algorithms using GES and MMHC in Step 1) are significantly worse than the oracle. Figs.~\ref{fig:result002P} and \ref{fig:result002H} also provides that the MRS algorithm is more accurate than state-of-the-art ODS, GES and MMHC algorithms in both precision and recall. It makes sense because the moments ratio scores the MRS algorithm exploits are less sensitive to the magnitude of the moments than the score the ODS algorithm uses as discussed in Section~\ref{SecStat}, and because the GES and MMHC algorithms recover up to the MEC by leaving some arrows undirected. However it must be pointed out that our MRS algorithm apply to GHD DAG models while GES and MMHC apply to general classes of DAG models. 

\begin{figure*}[!t]
	\centering \hspace{-5mm}
	\begin{subfigure}[!htb]{.23\textwidth} 
		\includegraphics[width=\textwidth]{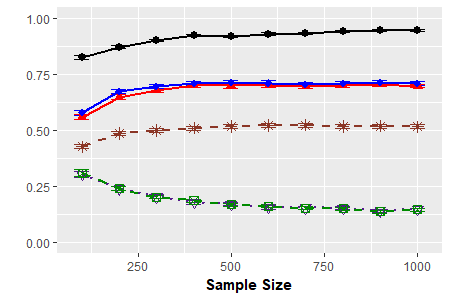}
		\caption{DAG Precision:$p=20$  }
	\end{subfigure} \hspace{-2mm}
	\begin{subfigure}[!htb]{.23\textwidth}
		\includegraphics[width=\textwidth]{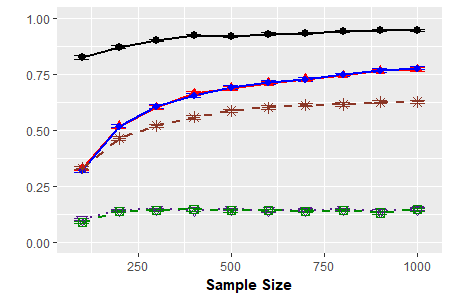}
		\caption{DAG Recall:$p=20$}
	\end{subfigure} \hspace{-2mm}
	\begin{subfigure}[!htb]{.23\textwidth}
		\includegraphics[width=\textwidth]{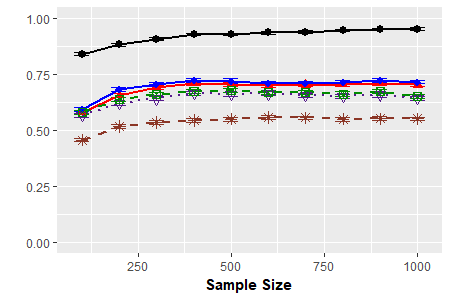}
		\caption{MEC Precision:$p=20$}
	\end{subfigure} \hspace{-2mm}
	\begin{subfigure}[!htb]{.23\textwidth}
		\includegraphics[width=\textwidth]{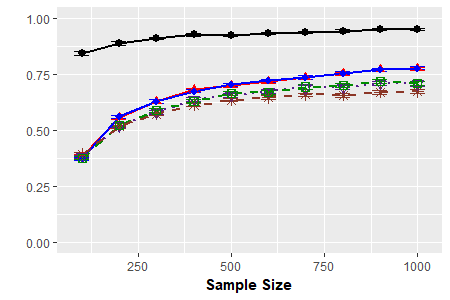}
		\caption{MEC Recall:$p=20$}
	\end{subfigure} \hspace{-2mm}
	\begin{subfigure}[!htb]{.07\textwidth}
		\includegraphics[width=\textwidth, trim={13cm 0 0cm 1.3cm},clip]{plots/PoissonP20Legend.png}
	\end{subfigure}
	\caption{Comparison of our MRS algorithms using GES and MMHC algorithms in Step 1) and $r = 2$ to the ODS, GES, MMHC algorithms in terms of precision and recall for Poisson DAG models with $p=20$.} 
	\label{fig:result005}
\end{figure*} 

Not surprisingly, the MRS algorithm performs better than ODS, GES and MMHC algorithms in terms of recovering MECs since the ordering is well estimated by the MRS algorithm. Fig.~\ref{fig:result005} empirically confirms this in Poisson DAG models with $p = 20$. 

\subsubsection{Computational Complexity}

\begin{figure}[!t]
	\centering \hspace{-2mm}
	\begin{subfigure}[!htb]{.35\textwidth} 
		\includegraphics[width=\textwidth, height=3cm]{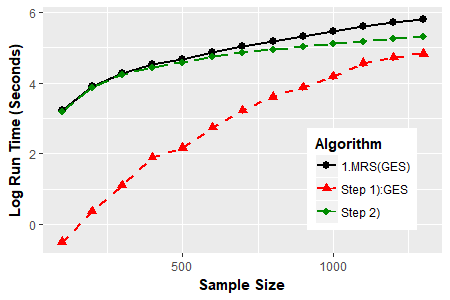}
		\caption{Varying $n$}
	\end{subfigure}
	\hspace{0.5cm}
	\begin{subfigure}[!htb]{.35\textwidth}
		\includegraphics[width=\textwidth, height=3cm]{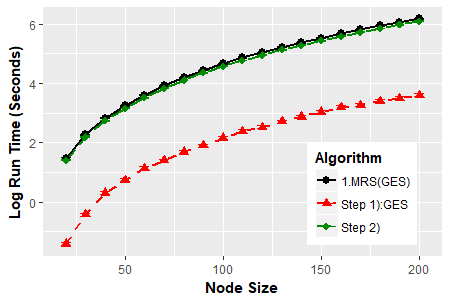}
		\caption{Varying $p$}
	\end{subfigure}
	\caption{Log run-time of the MRS algorithm using GES algorithm in Step 1) for learning Poisson DAG models with respect to (a) $n \in \{100, 200, ..., 1300\}$ with the fixed $p = 100$, and (b) $p \in \{10, 20, ..., 200\}$ with the fixed $n = 500$. } 
	\label{fig:result3} 
\end{figure}

 To validate the computational complexity discussed in Section~\ref{SecComp}, we show the log run-time of Step 1) and Step 2) of the MRS algorithm in Fig.~\ref{fig:result3} where the GES is applied for Step 1). We measured the run-time for learning Poisson DAG models by varying (a) $n \in \{100, 200, ..., 1300\}$ with the fixed node size $p= 100$, and (b) $p \in \{10, 20, ..., 200\}$ with the fixed sample size $n = 500$. As we see in Fig.~\ref{fig:result3}, the time complexity of Step 1) is $O(n^2 p^2)$, and that of Step 2) of the MRS algorithm is $O(n p^2)$. Hence we confirm the addition of estimation of ordering does not significantly add to the computational bottleneck.

\subsubsection{Deviations to the Distribution Assumptions}

\begin{figure}[!t] 
	\centering \hspace{-2mm}
	\begin{subfigure}[!htb]{.23\textwidth} \hspace{-0mm}
		\includegraphics[width=\textwidth, trim={8.4mm 0 0cm 0cm},clip]{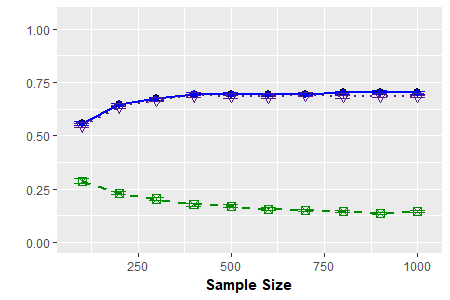}
		\caption{Poisson: Precision}
	\end{subfigure} 
	\begin{subfigure}[!htb]{.23\textwidth} 
		\includegraphics[width=\textwidth, trim={8.4mm 0 0cm 0cm},clip]{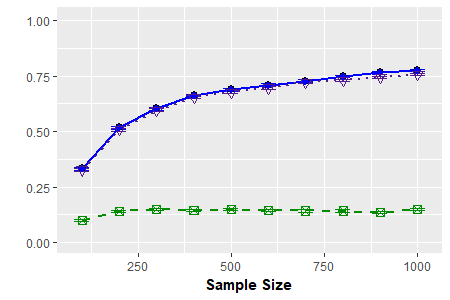}
		\caption{Poisson: Recall}
	\end{subfigure} \hspace{-2.8mm}
	\begin{subfigure}[!htb]{.23\textwidth} \hspace{-0mm}
		\includegraphics[width=\textwidth, trim={8.4mm 0 0cm 0cm},clip]{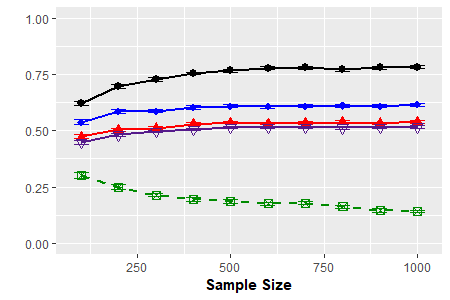}
		\caption{Hybrid: Precision}
	\end{subfigure} 
	\begin{subfigure}[!htb]{.23\textwidth} 
		\includegraphics[width=\textwidth, trim={8.4mm 0 0cm 0cm},clip]{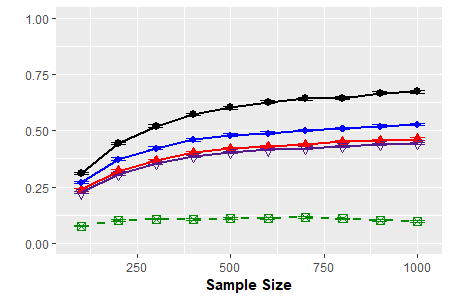}
		\caption{Hybrid: Recall}
	\end{subfigure} \hspace{-2.8mm}
	\begin{subfigure}[!htb]{.07\textwidth}
		\includegraphics[width=\textwidth, trim={13.5cm 0 0cm 1.3cm},clip]{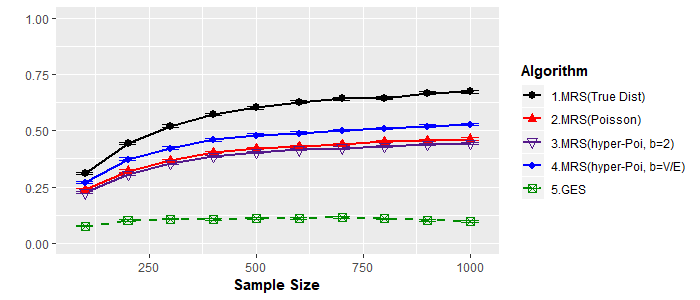}
	\end{subfigure}
	\caption{Comparison of the MRS algorithms with the different assumed node conditional distribution and the GES algorithm in terms of recovering Poisson and Hybrid DAG models with $p=20$. } 
	\label{fig:result4}
\end{figure}

 When the data are generated by a GHD DAG model where the conditional distribution of each node given its parents is unknown, our algorithm is not guaranteed to estimate the true graph and its ordering. Therefore, an important question is how well the MRS algorithm recovers graphs when incorrect distributions are used. In this section, we heuristically investigate this question. 

We use the same setting of the data generation for Poisson and Hybrid GHD DAG models with the node size $p =20$. We consider (i) the true (conditional) distributions, and assume all nodes (conditional) distributions are either (ii) Poisson; (iii) hyper-Poisson with $b = 2$; or (iv) hyper-Poisson with $b = \widehat{\var}(X)/\widehat{\E}(X)$ that is an estimator for the hyper-Poisson parameter $b$. We compare the MRS and GES algorithms by varying sample size $n \in \{100,200,...,1000\}$ in Fig.~\ref{fig:result4}.

Fig.~\ref{fig:result4} shows that the MRS algorithms recover the true graph better as sample size increases although there is no theoretical guarantees. It shows that the MRS algorithm enables to learn a part of ordering even if the true (conditional) distributions are unknown as long as there are sufficient samples.

\subsection{Real Multi-variate Count Data: 2009/2010 NBA Player Statistics}

\begin{figure}
	\centering
	\includegraphics[width=0.57\linewidth]{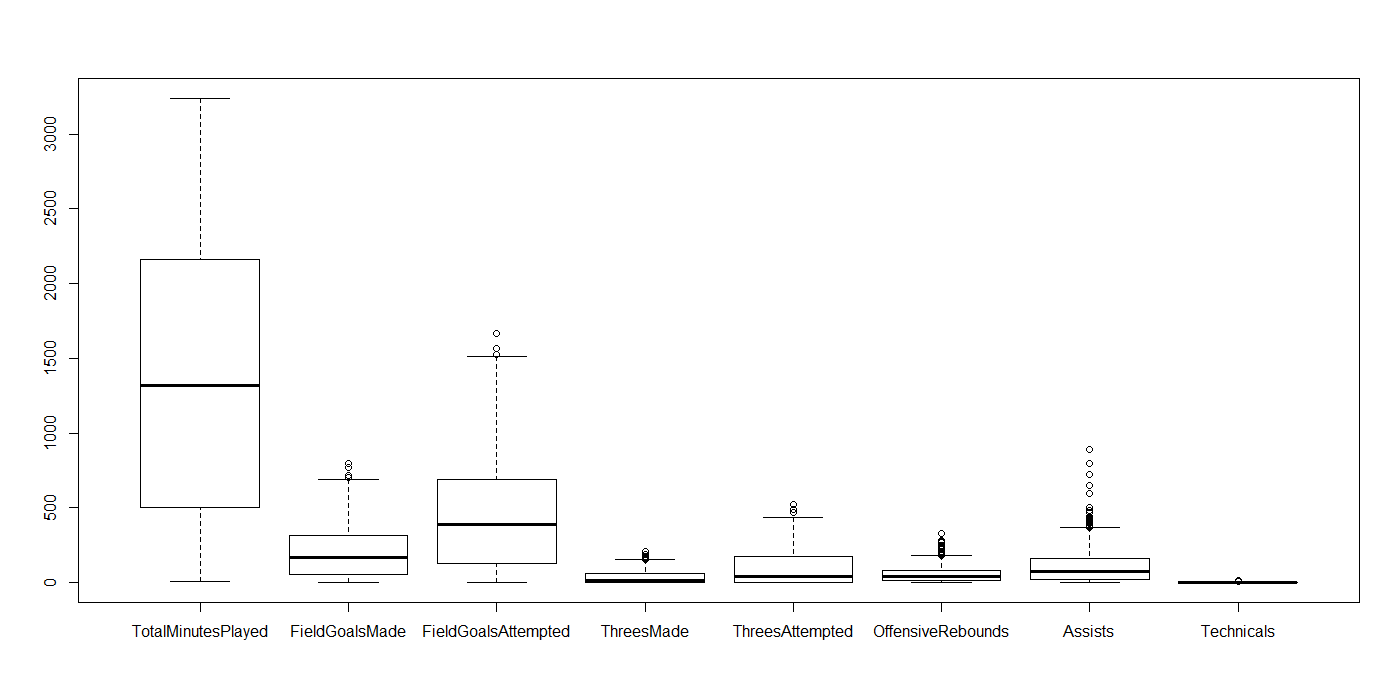}
	\includegraphics[width=0.4\linewidth]{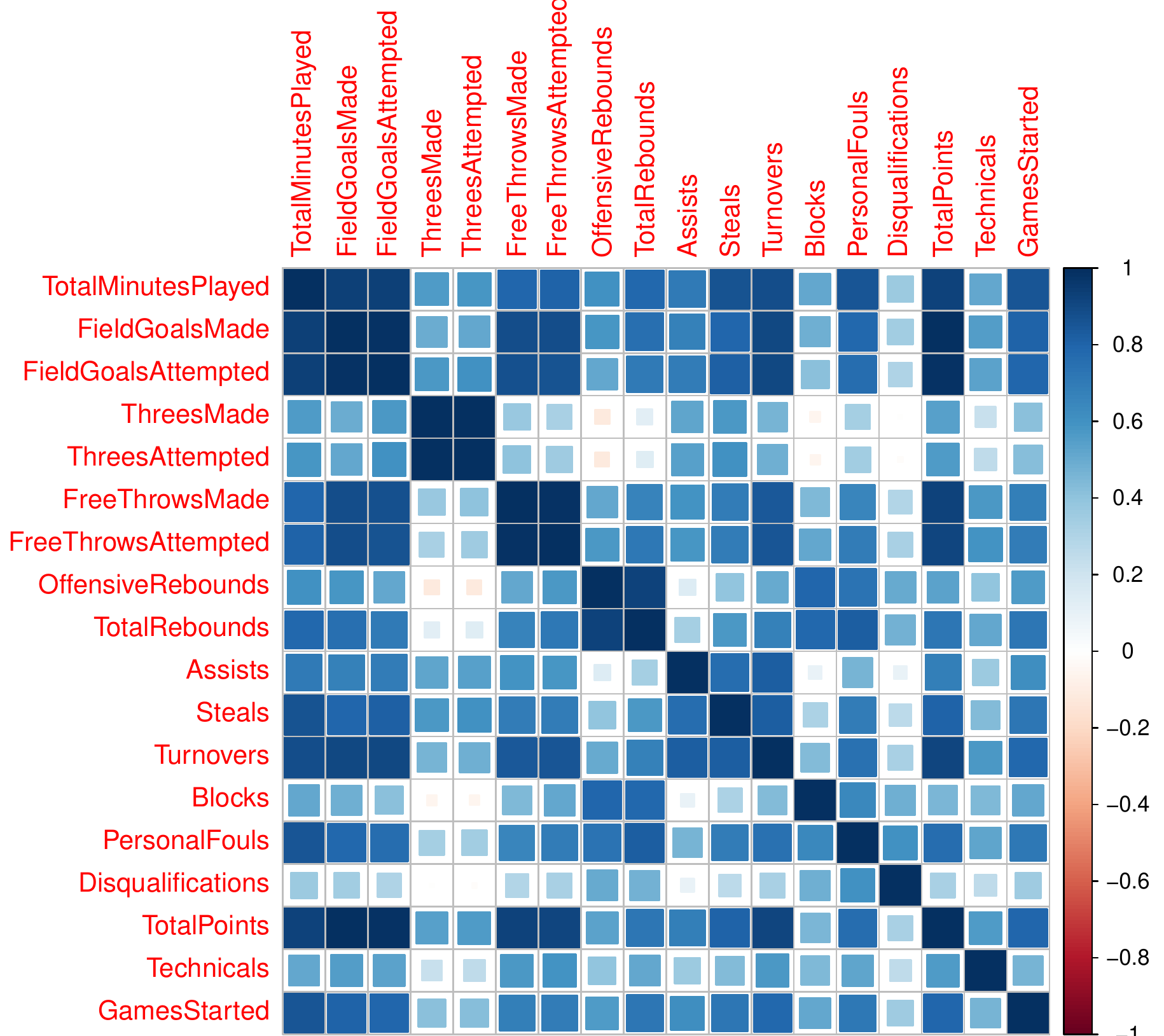}
	\caption{Box plots for some NBA statistics depending on positions (left). Box plots consider the total minutes played, total number of field goals made and attempted, threes made and attempted, offensive rebounds, assists, and technical fouls. Correlation Plots for NBA statistics (right). Blue represents a high correlation and white represents a small correlation. }
	\label{fig:nbaplot}  \vspace{-0.3cm}
\end{figure}

\label{SecReal}

We demonstrate the advantages of our graphical models for count-valued data by learning a 441 NBA player statistics from season 2009/2010 (see R package SportsAnalytics for detailed information) because we believe that some basketball statistics have causal or directional relationships. The original data set contains 24 covariates: player name, team name, player’s position, total minutes played, total number of field goals made, field goals attempted, threes made, threes attempted, free throws made, free throws attempted, offensive rebounds, rebounds, assists, steals, turnovers, blocks, personal fouls, disqualifications, technicals fouls, ejections, flagrant fouls, games started and total points. We eliminated player name, team name, number of games played, and player’s position, because our focus is to find the directional or causal relationships between statistics. We also eliminated ejections and flagrant fouls because both did not occur in our data set. Therefore the data set we consider contains 18 discrete variables. 

Fig.~\ref{fig:nbaplot} (left) shows that the magnitude of NBA statistics are significantly different, and hence we expect our MRS algorithm would be more accurate than the comparison ODS algorithm. Moreover, Fig.~\ref{fig:nbaplot} (right) shows that all 18 variables are positively correlated. This makes sense because the total minutes played is likely to be positively correlated with other statistics, and some statistics have causal or directional relationships (e.g., the more shooting attempt implies the more shooting made).


The MRS and ODS algorithms are applied where GES algorithm is used in Step 1). We assume that the conditional distribution of each node given its parents is hyper-Poisson because most of NBA statistics we consider are the number of successes or attempts counted in the season. We emphasize that our method requires a known conditional distribution assumption to recover the true graph. However since we do not have prior node distribution information, we set $b_j = \widehat{\var}(X_j)/\widehat{\E}(X_j)$ as we used in simulations that enables the MRS algorithm successfully recovers the directed edges.


\begin{figure}[!t]
	\centering \hspace{-2mm}
	\begin{subfigure}[!htb]{.4\textwidth}
		\includegraphics[width=\textwidth]{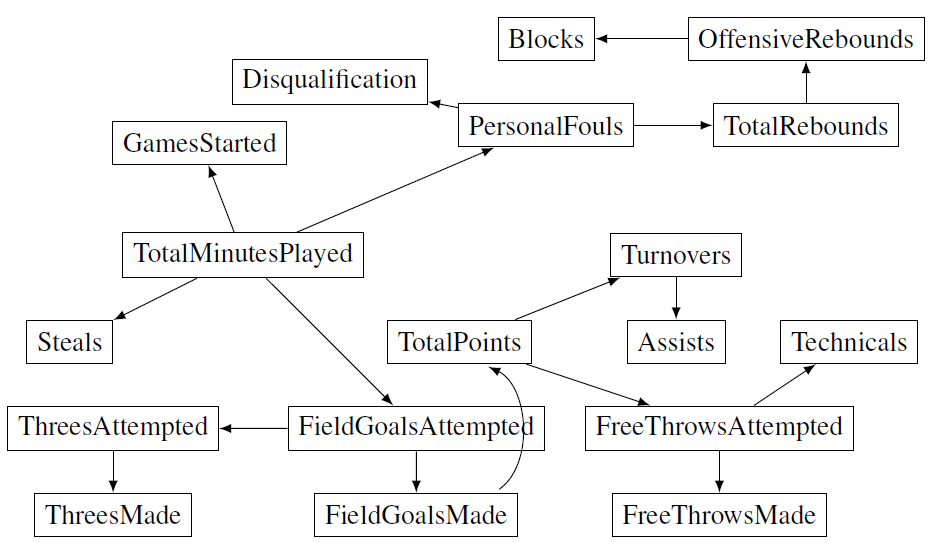}
		\caption{DAG from MRS }
	\end{subfigure}
	\begin{subfigure}[!htb]{.4\textwidth}
		\includegraphics[width=\textwidth]{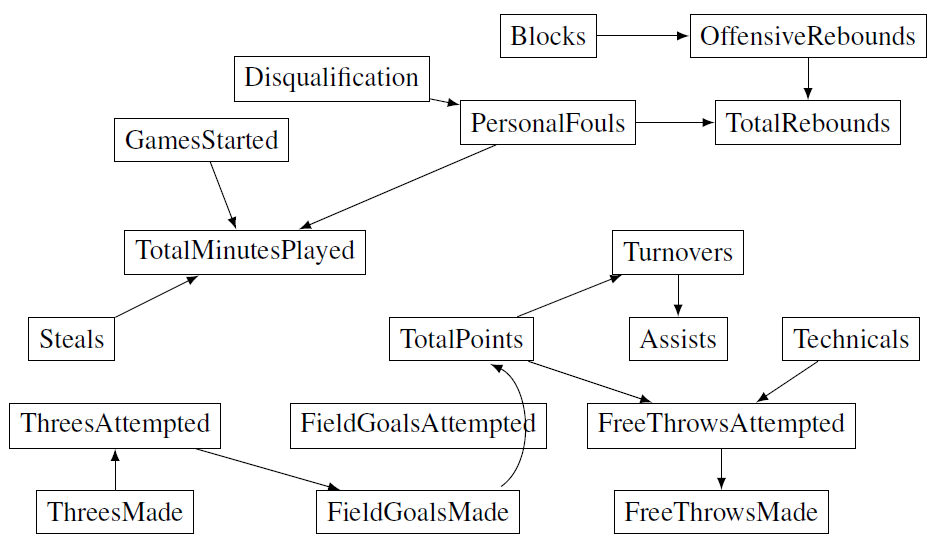}
		\caption{DAG from ODS }
	\end{subfigure}
	\caption{NBA players statistics directed graph estimated by by MRS (left) and ODS (right) algorithms.} 
	\label{fig:NBADAG} 
\end{figure}

	\begin{table}[!t]
		\begin{center}
			
		\begin{tabular}{l  l}
			\toprule
			Explanable edges &
			TotalMinutesPlayed $\rightarrow$ PersonalFouls, Steals and GamesStarted, \\
			&  ThreeAttempted $\rightarrow$ ThreeMade,
			TotalRebounds $\rightarrow$ OffensiveRebounds, \\
			& PersonalFouls $\rightarrow$ Disqualification \\
			\midrule
			Unexplanable edges &
			OffensiveRebounds $\rightarrow$ Blocks, FreeThrowsAttempted $\rightarrow$ Techincals \\
			\bottomrule
		\end{tabular}
		\caption{The set of directed edges in the estimated DAG via the MRS algorithm while the estimated DAG via the ODS algorithm has opposite directions. }
		\label{table:diffDAG}    \vspace{-1cm}
	\end{center}
	\end{table}

Fig.~\ref{fig:NBADAG} show the estimated directed graph using the MRS and ODS algorithm. We provide the differences between the estimated DAGs in Table~\ref{table:diffDAG}. In Table.~\ref{table:diffDAG}, we provide 8 distinct directed edges in the estimated DAG from the MRS algorithm while the estimated DAG from the ODS algorithm has opposite directions.

Explainable edges in Table.~\ref{table:diffDAG} shows the directed edges in the estimated DAG from the MRS algorithm while the estimated DAG from the ODS algorithm has opposite directions. This set of directed edges is more acceptable because the total minutes played  would be a reason for other statistics, and a large number of shooting attempted would lead to the more shootings made. It is consistent to our main point that MRS algorithm provides more legitimate directed edges than the ODS algorithm by allowing a broader class of count distributions. Unexplainable edges in Table.~\ref{table:diffDAG} shows the set of unaccountable edges in terms of causal or directional relationships regardless of directions. Hence they may be introduced by Step 1) estimation of the skeleton.

\section{Acknowledgments}

This work was supported by the National Research Foundation of Korea(NRF) grant funded by the Korea government(MSIT) (NRF-2018R1C1B5085420).

%


\bibliographystyle{plain} 

\bibliography{FMDAG_reference}

\clearpage

\section{Appendix}

\subsection{Proof for Proposition~\ref{prop:factorial}}

\label{SecProp1}

\begin{proof}
	For any positive integer $r \geq 1$, \cite{kemp1968wide} shows that 
	\begin{eqnarray}
	\label{eq:factorialmoment}
	\E((X_j)_r \mid X_{\pa(j)} ) 
	&=& \theta^r \prod_{i = 1}^{p} \frac{ (a_i+r-1)! }{ (a_i-1)! }  \prod_{j = 1}^{q} \frac{ (b_j-1)! }{ (b_j+r-1)!  } .
	\end{eqnarray}
	
	Then, the expectation can be obtained when $r = 1$. 
	\begin{eqnarray*}
		\label{eq:expectation}
		\E( X_j \mid X_{\pa(j)} ) 
		&=& \theta \times \prod_{i = 1}^{p} a_i \prod_{j = 1}^{q} \frac{1}{b_j}.
	\end{eqnarray*}
	
	By plugging this into ~Eqn. \eqref{eq:factorialmoment}, we have
	\begin{eqnarray*}
		\E((X_j)_r \mid X_{\pa(j)} ) 
		&=& \E(X_j \mid X_{\pa(j)} )^r \prod_{i = 1}^{p} \frac{ (a_i+r-1)! }{ (a_i-1)! a_i^r } \prod_{j = 1}^{q} \frac{ (b_j-1)! b_j^r }{ (b_j+r-1)!  }.
	\end{eqnarray*}
	
\end{proof}

\subsection{Proof for Theorem~\ref{Thm1}}

\label{SecThmCausalOrderingProof}

\begin{proof}
	
	Without loss of generality, we assume the true ordering is unique and $\pi = (\pi_1,...,\pi_p)$. For notational convenience, we define $X_{1:j} = (X_{\pi_1},X_{\pi_2},\cdots,X_{\pi_j})$ and $X_{1:0} = \emptyset$. We prove the  identifiability of our GHD DAG models by mathematical induction. 
	
	As we discussed in the main body of the paper, for any node $j \in V \setminus \{\pi_1\}$, Prop.~2.2 and Assumption~2.3 ensure that $$\frac{ \E( (X_j)_r ) }{ f_j^{(r)}(\E( X_j)) } > 1,  \text{ and } \quad \frac{ \E( (X_{\pi_1})_r ) }{ f_{\pi_1}^{(r)}(\E( X_{\pi_1})) } =  \frac{ f_{\pi_1}^{(r)}(\E( X_{\pi_1})) ) }{ f_{\pi_1}^{(r)}(\E( X_{\pi_1})) } =  1. 
	$$ 
	Hence we can determine $\pi_1$ as the first element of the ordering.
	
	For the $(m-1)^{th}$ element of the ordering, assume that the first $m-1$ elements of the ordering and their parents are correctly estimated. Now, we consider the $m^{th}$ element of the ordering and its parents. 
	Again Prop.~2.2 and Assumption~2.3 yield that for$j \in \{ \pi_{m+1},\pi_{m+2},\cdots, \pi_{p} \}$, 
	$$
	\frac{ \E( (X_j)_r)  }{ \E( f_j^{(r)}(\E( X_j \mid X_{1:(m-1)} ) ) ) }  > \frac{ \E( \E( (X_j)_r \mid X_{1:(m-1)} ) ) }{ \E(\E( f_j^{(r)}(\E( X_j \mid X_{\pa(j)} ) ) \mid X_{1:(m-1)} ))  } =  1.
	$$
	In addition, it is clear that 
	$$\frac{ \E( \E( (X_{\pi_m})_r \mid X_{1:(m-1)} ) ) }{\E( f_{\pi_m}^{(r)}(\E( X_{\pi_m} \mid X_{1:(m-1)} ) ) ) } = \frac{ \E( \E( (X_{\pi_m})_r \mid X_{\pa(\pi_m)} ) ) }{ \E( f_{\pi_m}^{(r)}(\E( X_{\pi_m} \mid X_{\pa(\pi_m) } ) ) ) } = \frac{ \E( (X_{\pi_m})_r ) }{ \E( f_{\pi_m}^{(r)}(\E( X_{\pi_m} \mid X_{\pa(\pi_m) } ) ) ) }  = 1.$$
	Hence we can estimate a valid $m^{th}$ component of the ordering $\pi_m$ and its parents by testing whether the r-th moments ratio is whether greater than or equal to 1. By the mathematical induction this completes the proof. 
	
\end{proof}

\subsection{Proof for Theorem~\ref{ThmCausalOrdering}}

\label{SecThmCausalOrderingProof2}

\begin{proof}
	We first reintroduce some necessary notations and definitions to make the proof concise. Without loss of generality, assume that the true ordering is unique and $\pi = (\pi_1, ..., \pi_p)$. In addition, we assume the true skeleton is provided. For ease of notation, we drop the $r$ in the both r-th moments ratio scores and CMR function. Then, the element of the score can be written as:
	\begin{align*}
	\mathcal{S}(j,k)(X_{C_{jk}})  &:=\frac{ \E ( X_k^r \mid X_{C_{jk}} ) }{ f_k  ( \E( X_k \mid X_{C_{jk}}  )  ) - \sum_{m=0}^{r-1} s(r,m) \E(X_k^m \mid X_{C_{jk} } )  }, \mbox{ and} \\
	\S(j, k)(X_{\widehat{C}_{jk}})  &:=\frac{ \widehat{\E}( X_k^r \mid X_{ \widehat{C}_{jk}} ) }{ f_k ( \widehat{\E}( X_k \mid X_{\widehat{C}_{jk}}  )  ) - \sum_{m=0}^{r-1} s(r,m) \widehat{\E}(X_k^m \mid X_{\widehat{C}_{jk}} )  }.
	\end{align*}
	where $C_{jk}$ is the candidate parents set and $s(r,k)$ is Stirling numbers of the first kind. Hence the r-th moments ratio score is 
	\begin{eqnarray*}
		\widehat{\mathcal{S}}(j,k) :=  \sum_{x \in \mathcal{X}_{\widehat{C}_{jk}} } \frac{ n(x) }{ n_{\widehat{C}_{jk}} }\widehat{\mathcal{S}}_r(j,k)(x).
	\end{eqnarray*}
	
	We define the following important events: For each node $j \in V$ and set $S \subset V \setminus (\de(j) \cup \{j\})$ and for any $\epsilon > 0$, let
	\begin{eqnarray}
	\label{set:zeta}
	\zeta_1 & := & \left\{ \min_{j = 1, ...,p-1} \min_{k = j,...,p} \left| \mathcal{S}(j,\pi_k) - \S(j,\pi_k) \right| > \frac{M_{\min}}{2}  \right\} \nonumber \\
	\zeta_2 & := & \left\{ \max_{j \in V} \left| \widehat{\E}( X_j^r \mid X_S)  - \E( X_j^r \mid X_S) \right| < \epsilon \right\}  \nonumber \\
	\zeta_3 & := & \left\{ \max_{j \in V} \left| f_j  \left( \widehat{\E}(X_j \mid X_S)  \right) - f_j  \left( \E( X_j \mid X_S ) \right) \right| < \frac{ \epsilon }{ 2 }  \right\}  \nonumber \\
	\zeta_4 & := & \left\{ \max_{j \in V} \left| \left(  \sum_{k=0}^{r-1} s(r,k) \widehat{\E}(X_j^k\mid X_{S})  - \sum_{k=0}^{r-1} s(r,k)\E(X_j^k\mid X_{S}) \right) \right| < \frac{ \epsilon }{ 2 }  \right\}  \nonumber \\	
	\zeta_5 & := & \left\{ \max_{j \in V}  \max_{i \in \{ 1,2,\cdots,n\} } |X_j^{(i)}| < 4 \log \eta \right\}.
	\end{eqnarray}
	Here we use the method of moments estimators $\frac{1}{n} \sum_{i = 1}^{n} (X_j^{(i)})^k$ as unbiased estimators for  $\E( X_j^k )$ for all $ 1 \leq k \leq r$.
	
	We prove that our algorithm recovers the ordering of any GHD DAG model in the high dimensional settings if the indgree is bounded. The probability that ordering is correctly estimated from our method can be written as
	\begin{eqnarray*}
		& &P\left( \widehat{\pi} = \pi \right) \\
		& =&P\left( \S(1, \pi_1) < \min_{j = 2,...,p} \S(1, \pi_j),  \S(2, \pi_2) < \min_{j =3,...,p} \S(1, \pi_j), ..., \S(p-1, \pi_{p-1}) < \S(p-1, \pi_p) \right) \\
		& =& P\left( \min_{j = 1, ...,p-1} \min_{k = j+1,...,p} \S(j, \pi_k) -\S(j, \pi_j) > 0 \right)\\
		& =& P\left( \min_{j = 1, ...,p-1} \min_{k = j+1,...,p} \left\{ \left( \mathcal{S}(j, \pi_k) -\mathcal{S}(j, \pi_j) \right) - \left( \mathcal{S}(j,\pi_k) - \S(j,\pi_k) \right) + \left( \mathcal{S}(j,\pi_j) + \S(j,\pi_j) \right) \right\} > 0  \right) \\
		& \geq & P\left( \min_{j = 1, ...,p-1} \min_{k = j+1,...,p} \left\{ \left( \mathcal{S}(j, \pi_k) -\mathcal{S}(j, \pi_j) \right) \right\} > M_{\min} , ~and  \min_{j = 1, ...,p-1} \min_{k = j,...,p} \left| \mathcal{S}(j,\pi_k) - \S(j,\pi_k) \right| < \frac{M_{\min}}{2}  \right).
	\end{eqnarray*}
	
	By Assumption 3.1 (A1) $\mathcal{S}(j, \pi_k)  > 1 + M_{\min}$, the above lower bound of the probability is reduced to
	\begin{eqnarray*}
		\label{eqn:mainlowerbound}
		P\left( \widehat{\pi} = \pi \right) 
		&\geq& 1- P\left( \min_{j = 1, ...,p-1} \min_{k = j,...,p} \left| \mathcal{S}(j,\pi_k) - \S(j,\pi_k) \right| > \frac{M_{\min}}{2}  \right) \\
		&= & 1 - P(\zeta_1) \\
		&= & 1 - \left\{ P(\zeta_1 \mid \zeta_2, \zeta_3, \zeta_4) P(\zeta_2, \zeta_3, \zeta_4) + P(\zeta_1 \mid ( \zeta_2,\zeta_3,\zeta_4 )^c ) P( (\zeta_2, \zeta_3,\zeta_4)^c ) \right\} \\
		&\geq & 1 - \left\{ P(\zeta_1 \mid \zeta_2, \zeta_3, \zeta_4) + P( (\zeta_2, \zeta_3, \zeta_4)^c ) \right\} \\
		&= & 1 - \left\{ P(\zeta_1 \mid \zeta_2, \zeta_3, \zeta_4) + P( (\zeta_2, \zeta_3, \zeta_4)^c \mid \zeta_5) P(\zeta_5) + P( (\zeta_2, \zeta_3, \zeta_4)^c \mid \zeta_5^c ) P(\zeta_5^c) \right\} \\
		&\geq & 1 - \left\{ P(\zeta_1 \mid \zeta_2, \zeta_3, \zeta_4) + P( (\zeta_2, \zeta_3, \zeta_4)^c \mid \zeta_5) + P(\zeta_5^c) \right\} \\
		&\geq & 1 - \Big\{ \underbrace{ P(\zeta_1 \mid \zeta_2, \zeta_3, \zeta_4) }_{Prop\ref{prop1}} + \underbrace{ P(\zeta_2^c \mid \zeta_5 ) + P(\zeta_3^c \mid \zeta_5 )
			+ P(\zeta_4^c \mid \zeta_5 ) }_{Prop\ref{prop2}}+ \underbrace{ P(\zeta_5^c) }_{Prop\ref{prop3}} \Big\}. \\
	\end{eqnarray*}
	
	Next we introduce the following three propositions to show that the above lower bound converges to 1. The first proposition proves that estimated score is accurate under some regularity conditions. For ease of notation, let
	\begin{align*}
	g_j( \widehat{\E}( X_j \mid X_{\widehat{C}_{jk}} ) ) =  f_k  ( \widehat{\E}( X_k \mid X_{\widehat{C}_{jk}}  )  ) - \sum_{m=0}^{r-1} s(r,m) \widehat{\E}(X_k^m \mid X_{\widehat{C}_{jk}} ) .
	\end{align*}
	
	\begin{proposition}
		\label{prop1}
		Given the sets $\zeta_2, \zeta_3, \zeta_4$ in Egn.~\eqref{set:zeta}, $P(\zeta_1 \mid \zeta_2, \zeta_3, \zeta_4) = 0$ if one of the following conditions are satisfied for any $S \subset V \setminus (\de(j) \cup \{j\})$:
		%
		\begin{itemize}
			\item[(i)] $2 \E( X_j^r \mid X_S ) + ( 2 - M_{\min} ) g_j( \E( X_j \mid X_S ) ) \leq 0$ or 
			\item[(ii)] $\epsilon < \frac{ M_{\min} g_j( \E( X_j \mid X_S ) )^2 }{ \left(  2 \E( X_j^r \mid X_S ) + ( 2 - M_{\min} ) g_j( \E( X_j \mid X_S ) )  \right) }$.
		\end{itemize}
		
	\end{proposition}
	The first condition (i) is satisfied if $M_{\min}$ in Assumption 3.1 (A1) is sufficiently large and the second condition (ii) is satisfied if $\epsilon$ is sufficiently small. This means that if the estimated  r-th factorial moment is close to the true r-th factorial moment, then $\zeta_1$ is not satisfied with probability 1. Hence we discuss the error bound for the r-th factorial moment estimator in the next. 
	
	The following propositions show the error bound for the higher order moment $X_j^k$ for $1 \leq k \leq r$ given the set $\zeta_5$, and therefore the error bound for the r-th factorial moment estimator:
	\begin{proposition}
		\label{prop2}
		For any node $j \in V$ and any set $S \subset V \setminus (\de(j) \cup \{j\})$ and for any $\epsilon >0$,
		\begin{itemize}
			\item[(i)] For $\zeta_2$, 
			\begin{eqnarray*} 
				P(\zeta_2^c \mid \zeta_5) \leq 2\cdot p \cdot \exp \left\{ -\frac{  2 N_{\min} \epsilon^2 }{ (4 \log^2 \eta)^r } \right\}.
			\end{eqnarray*}
			\item[(ii)] For $\zeta_3$, and $m \in (\E(X_j \mid X_S) - \epsilon/2, \E(X_j \mid X_S) + \epsilon/2)$, 
			\begin{eqnarray*}
				P(\zeta_3^c \mid \zeta_5) \leq 2\cdot p \cdot\exp \left\{ - \frac{ N_{\min} \epsilon^2 }{ 8 ( \max( f_j'(m) ) )^2 \log^2 \eta } \right\}.
			\end{eqnarray*}
			\item[(iii)] For $\zeta_4$, 
			\begin{eqnarray*}
				P(\zeta_4^c \mid \zeta_5) \leq 2\cdot p\cdot r\cdot \exp  \left\{ -\frac{  2 N_{\min} \epsilon^2 }{  \max_{k \in \{1,...,r-1\}} s(r,k)  (4\log^{2} \eta)^r } \right\}.
			\end{eqnarray*}
		\end{itemize}
		where $N_{\min}$ is a predetermined minimum sample size in Assumption 3.1 (A3) and $s(r,k)$ is Stirling numbers of the first kind. 
	\end{proposition}
	
	\begin{proposition}
		\label{prop3}
		Under Assumption~3.1 (A2),
		\begin{eqnarray*}
			P(\zeta_5^c) \leq \frac{V_1}{\eta^2}.
		\end{eqnarray*}
	\end{proposition}

	Hence for any $\epsilon \in \left(0, \left| \frac{ M_{\min} g_j( \E( X_j \mid X_S ) )^2 }{ \left(  2 \E( (X_j)_r \mid X_S ) + ( 2 - M_{\min} ) g_j( \E( X_j \mid X_S ) ) \right) } \right| \right)$, the MRS algorithm recovers the true ordering at least of 
	\begin{eqnarray*}
		P\left( \widehat{\pi} = \pi \right) &\geq & 1 - \Big\{ \underbrace{ P(\zeta_1 \mid \zeta_2, \zeta_3, \zeta_4) }_{Prop\ref{prop1}} + \underbrace{ P(\zeta_2^c \mid \zeta_5 ) + P(\zeta_3^c \mid \zeta_5 ) + P(\zeta_4^c \mid \zeta_5 )}_{Prop\ref{prop2}}+ \underbrace{ P(\zeta_5^c) }_{Prop\ref{prop3}} \Big\}. \\
		& = & 1 -2\cdot p\cdot \exp \left\{ -\frac{  2 N_{\min} \epsilon^2 }{ (4 \log^2 \eta)^r } \right\}
		- 2 \cdot p\cdot \exp \left\{ - \frac{ N_{\min} \epsilon^2 }{ 8 (  \max( f_j'(m) ) )^2 \log^2 \eta } \right\} \\
		&& \quad - 2\cdot p\cdot r \cdot \exp \left\{ -\frac{  2 N_{\min} \epsilon^2 }{ \max_{k \in \{1,...,r-1\}} s(r,k) (4 \log^2 \eta)^r } \right\}
		- \frac{V_1}{\eta^2}. \\
	\end{eqnarray*}
	This result clams that if $N_{\min} = O( \log^{2r}(\eta) \log(p) )$, our algorithm correctly estimate the ordering of the graph. 
	
	Lastly, we show the relationship between the sample size $n$ and $N_{\min}$ to satisfy Assumption 3.1 (A3). Suppose that $d$ is the maximum number of parents of a node. Then the maximum size of the candidate parents set is $d$. The scenario is that a conditioning set has two possible cases. If there is only one element for the conditioning set, there is no difference between the conditional and marginal distributions. In the best case, $n = 2 N_{\min}$ when the there are two conditional distributions $|\mathcal{X}_{C}| = 2$. Hence if $n = O( (\log^{2r}(\eta)  ( \log(p) + \log(r) ) )$, our algorithm works in the high dimensional settings. In the worst case given $\zeta_5$, the sample size is $n = ( 4 \log(\eta)^d -2 )( N_{\min} -1 ) + 2 N_{\min} = 4 \log(\eta)^d (N_{\min} - 1) + 2 $ where the number of all elements of $\{x \in \mathcal{X}_{C} \mid \sum_{i}^{n} \mathbf{1}(X_{C}^{(i)} = x) \geq N_{\min} \}$ is two and all other elements of $\mathcal{X}_{C}$ has $N_{\min} -1$ repetitions. In this worst case, if $n = O( \log(\eta)^{(2r+d)} ( \log(p) + \log(r) ) )$ our algorithm correctly recovers the ordering with high probability. 
	
\end{proof}

\subsubsection{Proof for Proposition \ref{prop1} }
\begin{proof}
	For ease of notation, let $\eta = \max\{n, p\}$ and the r-th moments ratio score:
	\begin{eqnarray*}
		\widehat{\mathcal{S}}(j,k) :=  \sum_{x \in \mathcal{X}_{\widehat{C}_{jk}} } \frac{ n(x) }{ n_{\widehat{C}_{jk}} }\widehat{\mathcal{S}}_r(j,k)(x).
	\end{eqnarray*}
	In addition, let
	\begin{align*}
	g_k( \widehat{\E}( X_k \mid X_{\widehat{C}_{jk}} ) ) =  f_k  ( \widehat{\E}( X_k \mid X_{\widehat{C}_{jk}}  )  ) - \sum_{m=0}^{r-1} s(r,m) \widehat{\E}(X_k^m \mid X_{\widehat{C}_{jk}} ) .
	\end{align*}
	
	For any $j \in V$, $k \in \{ \pi_{j}, ..., \pi_{p} \}$ and $x \in \mathcal{X}_{\widehat{C}_{jk}}$, we have
	\begin{eqnarray*}
		& & P\left( |\S(j,k)(x) - \mathcal{S}(j,k)(x)| > \frac{ M_{min} }{ 2 } \Big\vert \zeta_2, \zeta_3, \zeta_4 \right) \\
		& = & P\left( \left| \frac{ \widehat{\E}( X_k^r \mid x ) }{ g_k( \widehat{\E}( X_k \mid x ) )} - \frac{ \E( X_k^r \mid x ) ) }{ g_k( \E( X_k \mid x ) )} \right| > \frac{ M_{min} }{ 2 }  \bigg\vert \zeta_2, \zeta_3, \zeta_4 \right) \\
		&\leq & P\left( \frac{ \E( X_k^r \mid x ) + \epsilon }{ g_k( \E( X_k \mid x ) ) - \epsilon} - \frac{ \E( X_k^r \mid x ) ) }{ g_k( \E( X_k \mid x ) )} > \frac{ M_{min} }{ 2 } ~or~ \frac{\E( X_k^r \mid x ) ) }{ g_k( \E( X_k \mid x ) )} - \frac{ \E( X_k^r \mid x ) - \epsilon }{ g_k( \E( X_k \mid x ) ) + \epsilon}  > \frac{ M_{min} }{ 2 } \right) \\
		& = & P\left( \frac{ \epsilon ( g_k( \E( X_k \mid x ) ) +   \E( X_k^r \mid x )) }{ g_k( \E( X_k \mid x ) ) ( g_k( \E( X_k \mid x ) ) - \epsilon) } > \frac{M_{\min} }{2}  ~or~ \frac{ \epsilon ( g_k( \E( X_k \mid x ) ) +   \E( X_k^r \mid x )) }{ g_k( \E( X_k \mid x ) ) ( g_k( \E( X_k \mid x ) ) + \epsilon) } > \frac{M_{\min} }{2}   
		\right) \\
		& = & P\left( M_{\min} g_k( \E( X_k \mid x ) )^2  < \epsilon \left(  2 \E( X_k^r \mid x ) + ( 2 - M_{\min} ) g_k( \E( X_k \mid x ) )  \right)
		\right).
	\end{eqnarray*}
	
	Simple calculations yield that the above upper bound is zero if either 
	\begin{itemize}
		\item[(i)] $2 \E( X_k^r \mid x ) + ( 2 - M_{\min} ) g_k( \E( X_k \mid x ) ) \leq 0$ or 
		\item[(ii)] $\epsilon < \frac{ M_{\min} g_k( \E( X_k \mid x ) )^2 }{ \left(  2 \E( X_k^r \mid x ) + ( 2 - M_{\min} ) g_k( \E( X_k \mid x ) )  \right) }$.
	\end{itemize}
\end{proof}

\subsubsection{Proof for Proposition \ref{prop2}}
Since the proof for Prop. \ref{prop2} (i) - (iii) are analogous, we provide the proof for (iii) and then we provide the proof for (ii). 

\begin{proof}
	
	
	Using Hoeffding's inequality given $\zeta_5$, for $1 \leq k \leq r$ and any $\epsilon>0$,
	\begin{eqnarray*}
		P\left( \left| \widehat{\E}(X_j^k \mid X_S)  - \E( X_j^k \mid X_S ) \right| > \epsilon \right)  \leq 2\cdot p \cdot \exp \left\{ - \frac{  N_{\min} \epsilon^2 }{ 8 \log^{2k} \eta } \right\}.
	\end{eqnarray*}
	
	Hence, given $\zeta_5$, 
	\begin{eqnarray*}
		& & P\left( \left|  \sum_{k=0}^{r-1} s(r,k) \widehat{\E}(X_j^k\mid X_{S})  - \sum_{k=0}^{r-1} s(r,k)\E(X_j^k\mid X_{S}) \right| > \epsilon \mid \zeta_5 \right)  \\
		&\leq& \sum_{k =1}^{r-1} P\left( \left|  \widehat{\E}(X_j^k \mid X_S) - \E( X_j^k \mid X_S) \right| >  \frac{\epsilon}{ s(r,k) } \mid \zeta_5 \right) \\
		&\leq& \sum_{k =1}^{r-1} 2 \cdot p \cdot \exp \left\{ - \frac{  N_{\min} \epsilon^2 }{ 8 s(r,k) \log^{2k} \eta } \right\} \\
		&\leq& 2 \cdot p\cdot r \cdot \exp \left\{ - \frac{  N_{\min} \epsilon^2 }{ 8 \max_{k} s(r,k) \log^{2r} \eta } \right\}.
	\end{eqnarray*}
\end{proof}

Now we provide the proof for (ii).

\begin{proof}
	By Mean value theorem, we obtain
	\begin{equation*}
	f_j ( \widehat{\E}(X_j \mid X_S)  ) - f_j ( \E( X_j \mid X_S ) )  =  f_j'(\bar{m}) \left( \widehat{\E}(X_j \mid X_S) - \E( X_j \mid X_S ) \right). 
	\end{equation*}
	where $f_j'$ is the first derivative of $f_j$ and $\bar{m}$ is some point between $\widehat{\E}(X_j \mid X_S)$ and $\E( X_j  \mid X_S)$.
	
	Given the $| \widehat{\E}(X_j) - \E( X_j ) | < \epsilon/2$ from Prop.~\ref{prop2}(iii), we obtain
	\begin{equation*}
	f_j ( \widehat{\E}(X_j \mid X_S)  ) - f_j ( \E( X_j \mid X_S ) )  = \max_m{  f_j'(m) } \left( \widehat{\E}(X_j \mid X_S) - \E( X_j \mid X_S ) \right). 
	\end{equation*}
	for $m \in (\E( X_j \mid X_S ) - \epsilon/2, \E( X_j \mid X_S ) + \epsilon/2)$. Again applying Hoeffding's inequality given $\zeta_5$,  for any $\epsilon>0$,
	\begin{eqnarray*}
		P\left( \min_{j \in V} f_j  \left( \widehat{\E}(X_j \mid X_S)  \right) - f_j  \left( \E( X_j ) \right) > \epsilon \mid \zeta_5 \right)  
		&\leq& p\cdot \min_{j \in V} P\left( \left(  \widehat{\E}(X_j \mid X_S) - \E( X_j \mid X_S) \right) >  \frac{\epsilon}{ \max_m{  f_j'(m) } } \mid \zeta_5 \right) \\
		&\leq& 2\cdot p \cdot \exp \left\{ - \frac{  N_{\min} \epsilon^2 }{ 8 (\max_m{  f_j'(m) } )^2 \log^2 \eta } \right\}.
	\end{eqnarray*}
\end{proof}

\subsubsection{Proof for Proposition \ref{prop3}}
\begin{proof}
	The proof is directly from the concentration bound:
	\begin{eqnarray*}
		P(\zeta_5^c) 
		&=& P\left( \min_{j \in V} \min_{i \in \{1,2,...,n\} } \left| X_j^{(i)} \right| > 4 \log \eta \right) \\
		&\leq& n \cdot p\cdot P\left( P\left| X_j^{(i)} \right| > 4 \log \eta \right) \\
		&\leq& n\cdot p\cdot \frac{ \E( \exp(X_j^{(i)} )  ) }{\eta^4} \\
		&\leq& n\cdot p\cdot \frac{ \E( \E( \exp(X_j^{(i)} ) \mid X_{\pa(j)} ) ) }{\eta^4} \\				
		&\stackrel{(a)}{\leq}& n\cdot p\cdot \frac{V_1}{\eta^4} \\
		&\leq& \frac{V_1}{\eta^2}.
	\end{eqnarray*}
	Inequality (a) is from Assumption 3.1 (A2).
\end{proof}

\end{document}